\pdfoutput=1
\documentclass{article}

\PassOptionsToPackage{numbers, compress}{natbib}

\usepackage[main, final]{neurips_2026}

\usepackage[utf8]{inputenc} 
\usepackage[T1]{fontenc}    
\usepackage{hyperref}       
\usepackage{url}            
\usepackage{booktabs}       
\usepackage{amsfonts}       
\usepackage{nicefrac}       
\usepackage[table]{xcolor}  
\usepackage{xspace}
\usepackage{graphicx}
\usepackage{subcaption}
\usepackage{multicol}
\usepackage{multirow}
\usepackage{makecell}
\usepackage{amsmath}
\usepackage{pifont}
\usepackage{tcolorbox}
\usepackage{enumitem}
\usepackage{wrapfig}
\usepackage{algorithm}
\usepackage{algpseudocode}
\usepackage{algorithmicx}

\newcommand{\eg}{\textit{e.g.}\xspace}
\newcommand{\ie}{\textit{i.e.}\xspace}

\newcommand{\oursbenchmark}{{{PICL}}\xspace}
\newcommand{\ours}{{{ProCoRe}}\xspace}
\newcommand{\oursfull}{{{Proposition-level Contrastive Reward}}\xspace}

\title{Visual Grounding First,  Multimodal In-context Learning Follows}

\author{
    \textbf{Minhyuk Seo}$^{1,}$\thanks{indicates equal contribution. $^\dagger$ indicates corresponding authors. $~$  Work done while MS is with SNU and KU Leuven} \hspace{.35em}
    \textbf{Minjae Lee}$^{2,*}$ \hspace{.2em}
    \textbf{Chaeeun Lee}$^2$ \hspace{.2em}
    \textbf{Wei Lin}$^4$ \hspace{.2em}
    \textbf{Muhammad Jehanzeb Mirza}$^5$ \hspace{.2em} \\ \vspace{0.5em}
    \textbf{Tinne Tuytelaars}$^{3,\dagger}$ 
    \hspace{.2em}
    \textbf{Jonghyun Choi}$^{2,\dagger}$ \hspace{.2em}
    \\
    $^1$ UC Berkeley \quad
    $^2$ Seoul National University \quad
    $^3$ KU Leuven \quad
    $^4$ Xero \quad
    $^5$ MIT CSAIL \quad \\
}

\begin{document}

\maketitle

\begin{figure*}[h!]
    \centering   
    \includegraphics[width=\linewidth]{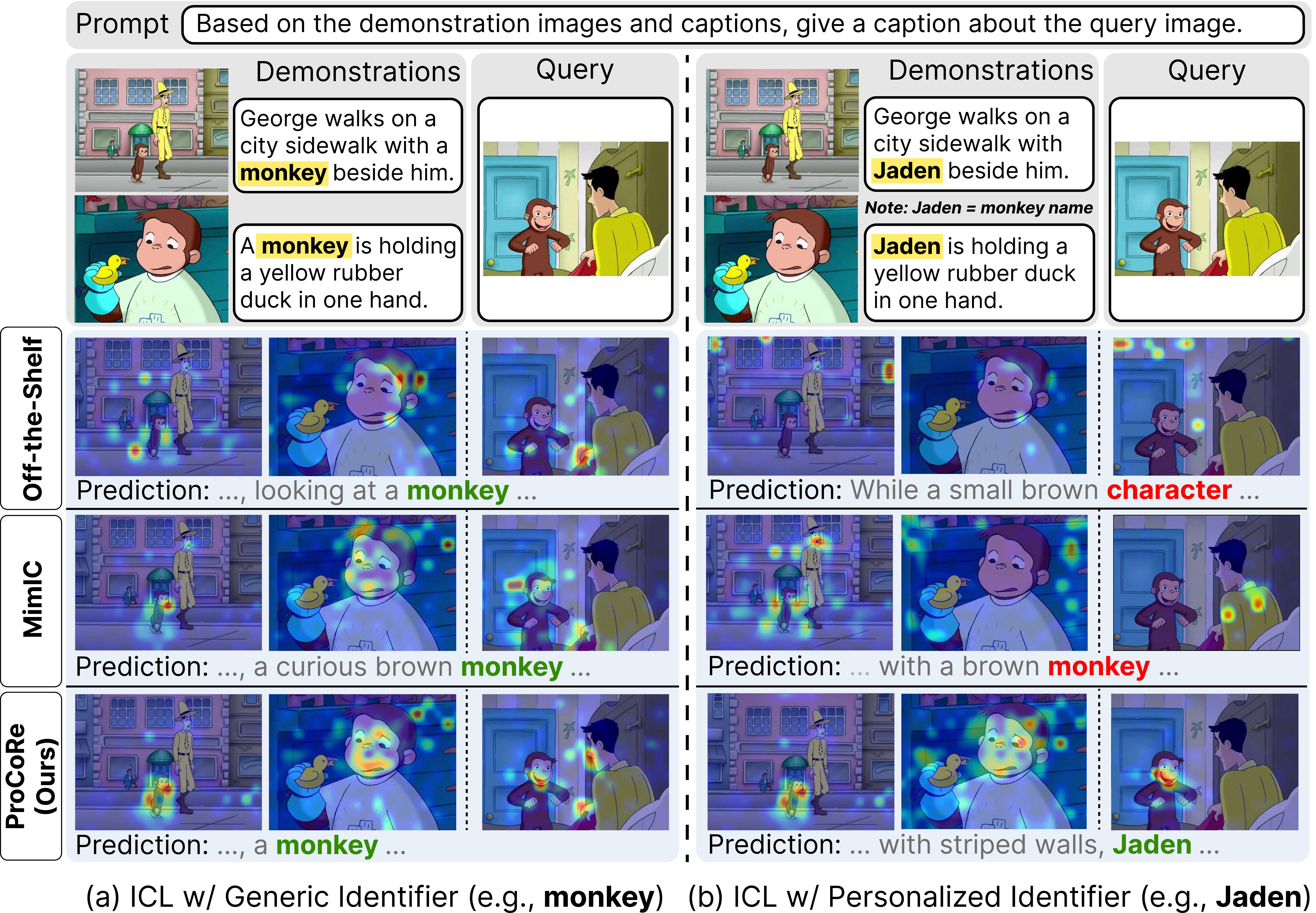}
    \caption{
        \textbf{Visual grounding in multimodal ICL with generic vs. personalized identifiers.}
        Attention maps show visual-token attention when predicting the highlighted token using Qwen-VL-3-8B.
        (a) When a generic identifier (\eg, ``{\setlength{\fboxsep}{0.5pt}%
\colorbox[HTML]{FFE32F}{\textbf{{monkey}}}}'') seen during pre-training appears in the ICL demonstrations, both baselines and \ours attend to visual regions relevant to the predicted token (\ie, \textbf{\textcolor[HTML]{318C00}{monkey}}).
        (b) In contrast, when the generic identifier is replaced with a personalized identifier (\eg, ``{\setlength{\fboxsep}{0.5pt}%
\colorbox[HTML]{FFE32F}{\textbf{{Jaden}}}}''), baselines fail to attend to the relevant region and thereby fall back to generic terms (\eg, ``\textbf{\textcolor[HTML]{FF0004}{character}}'', ``\textbf{\textcolor[HTML]{FF0004}{monkey}}''), whereas \ours correctly predicts the personalized concept (\ie, ``\textbf{\textcolor[HTML]{318C00}{Jaden}}'') with strong visual grounding. See Sec.~\ref{sec:appx:attention_quali} for more examples.}
    \label{fig:highlevel-teaser}
\end{figure*} 



    



\begin{abstract}
LLMs have shown strong in-context learning (ICL) capability, but extending it to Multimodal LLMs (MLLMs) remains challenging. 
Prior multimodal ICL methods often rely on ICL-specific datasets for additional training or task vector extraction, which can improve performance on the same ICL benchmark used for adaptation.
However, they do not generalize well to other benchmarks and often induce forgetting in previously well-solved tasks.
For this reason, we identify `visual grounding' as the key bottleneck in multimodal ICL; MLLMs often fail to attend to task-relevant visual evidence in demonstrations, instead over-relying on textual cues and producing hallucinated outputs. 
To address this, we propose \ours to enhance visual grounding through contrastive reinforcement learning on multimodal contrastive data for improving multimodal ICL.
\ours generates captions for contrastive image pairs, decomposes them into propositions, and optimizes contrastive rewards that encourage alignment with corresponding images while discouraging mismatched ones. 
Despite never seeing ICL-formatted training examples, \ours improves average multimodal ICL classification accuracy by 5.6\% and captioning ROUGE-L by 13.8\% over the off-the-shelf Qwen-VL-3-8B model, outperforming ICL-data-dependent state of the arts across benchmarks.
Finally, we introduce \oursbenchmark, a personalized multimodal ICL captioning benchmark that tests whether MLLMs genuinely understand visual demonstrations rather than relying on pre-trained semantic priors.
%
%
\end{abstract}

\section{Introduction}
In-context learning (ICL) enables models to adapt to new tasks from a few demonstrations without parameter updates~\cite{brown2020language, dong2024survey}. 
While this capability is well established in Large Language Models (LLMs), extending it to Multimodal LLMs (MLLMs) remains more challenging~\cite{yi2024bridge, jiang2025mimic}. 
The challenge is not simply that demonstrations contain images.
Effective multimodal ICL requires binding visual evidence to textual outputs in demonstrations and transferring this visual–linguistic association to the query. 
However, recent MLLMs often fail to fully leverage visual information in demonstrations, instead over-relying on textual cues and simply mimicking the text demonstrations~\cite{chen2025true}.

\begin{figure*}
    \centering   
    \vspace{-1em}
    \includegraphics[width=\linewidth]{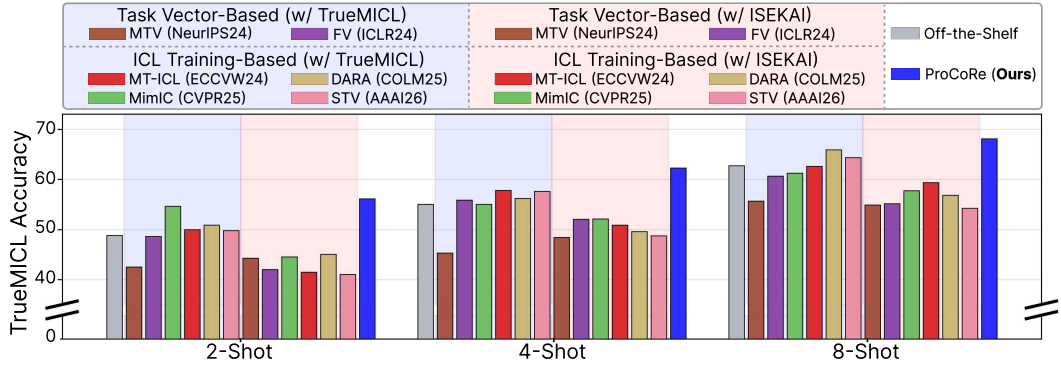}
    \vspace{-1.75em}
    \caption{
        \textbf{ICL data-dependent methods show limited generalization.}
        Blue-shaded bars denote ICL data-dependent baselines trained on TrueMICL ICL dataset, while red-shaded bars denote those based on ISEKAI.
        They often outperform the off-the-shelf model when evaluated on the same ICL dataset used for training (\ie, train: TrueMICL $\rightarrow$ test: TrueMICL), but degrade significantly when their training and test ICL datasets differ (\ie, train: ISEKAI $\rightarrow$ test: TrueMICL).
        In contrast, even without using any ICL data, \ours outperforms them through strong visual grounding.
    }
    \vspace{-0.5em}
    \label{fig:teaser}
\end{figure*}

Prior work has largely improved multimodal ICL by selecting useful demonstrations or adapting models to ICL-formatted data. 
Retrieval- and ordering-based methods~\cite{baldassini2024makes, luo2024does, chen2025can} retrieve and rank ICL demonstrations based on their similarity to the query.
Despite their effectiveness for ICL in LLMs, they are highly sensitive to prompt design and sample order in multimodal ICL, often underperforming even random selection~\cite{yang2024lever}. 
In contrast, ICL training-based methods tune MLLMs on ICL datasets~\cite{doveh2024towards, chen2025true, li2025m, jiang2025mimic}, while task vector-based methods~\cite{ma2026and, todd2024function} use ICL data to extract task vectors that steer hidden states, replicating the effect of demonstrations.
Despite their effectiveness, we observe that their reliance on ICL datasets limits their generalization (Fig.~\ref{fig:teaser}): they perform well on the same ICL benchmark used for training or task vector extraction, but drastically degrades on other ICL benchmarks.
Moreover, training on ICL datasets can cause catastrophic forgetting on previously well-solved tasks (\eg, simple VQA) (Fig.~\ref{fig:forgetting}). 
More fundamentally, since ICL in LLMs is widely considered an emergent capability from pre-training rather than explicit supervision on ICL-formatted datasets~\cite{lu2024emergent, muckatira2024emergent}, this raises a key question: \textit{do ICL data-dependent methods address the root cause of poor multimodal ICL, or merely improve performance on their training benchmarks?}


In this work, rather than relying on ICL benchmarks, we investigate a more fundamental limitation that hinders multimodal ICL by asking \textit{why multimodal ICL fails and what prevents MLLMs from effectively using demonstrations}. 
Prior work has shown that visual hallucination~\cite{wang2023evaluation, liu2024mitigating, gunjal2024detecting}, where outputs include nonexistent objects or inconsistent visual details, is a key factor behind poor multimodal ICL~\cite{shukor2024beyond}. 
With further analysis, we find that \textit{poor visual grounding}, a phenomenon attending to irrelevant regions instead of task-relevant visual evidence, significantly contributes to these hallucinations during multimodal ICL.
Specifically, we reveal that this issue is particularly pronounced for personalized identifiers unseen during pre-training: while MLLMs can ground correctly on generic identifiers common in training data (Fig.~\ref{fig:highlevel-teaser}-(a)), they struggle with personalized identifiers (Fig.~\ref{fig:highlevel-teaser}-(b)). 
This limitation is critical, as real-world multimodal ICL often involves personalized references, where demonstration captions include identifiers such as a dog named ``John'' rather than generic concepts like ``dog''~\cite{seifi2025personalization, seifi2026ego}. 
Since such concepts cannot be reliably inferred from pre-trained priors alone, correct predictions rely more directly on accurate visual grounding.

Motivated by this, we propose \ours (\textbf{Pro}position-level \textbf{Co}ntrastive \textbf{Re}ward), which improves multimodal ICL by enhancing visual grounding. 
Instead of relying on ICL datasets, \ours trains MLLMs to distinguish visually contrastive samples (\eg, Fig.~\ref{fig:contrastive_data_examples}), drawing inspiration from contrastive learning that fine-grained discrimination improves visual representation~\cite{chen2023protoclip}.
Specifically, it generates captions for contrastive image pairs, decomposes them into propositions (\ie, distinct propositions for each caption and common propositions), and applies contrastive reinforcement learning to align each proposition with its corresponding image while discouraging alignment with mismatched images.
By operating at the proposition-level rather than on sentence-level, \ours enables MLLMs to capture both distinctive and shared visual features.
As a result, \ours improves grounding for both generic and personalized concepts (Fig.~\ref{fig:highlevel-teaser}),  and significantly enhances multimodal ICL, despite using no ICL data for training (Fig.~\ref{fig:teaser}).
In addition, \ours mitigates forgetting on general VQA tasks (Fig.~\ref{fig:forgetting}).

Finally, to better assess whether MLLMs truly rely on multimodal demonstrations rather than pre-trained semantic priors, we introduce \oursbenchmark, a personalized ICL captioning benchmark built around personalized concept names. 
Existing evaluations often use common concepts such as ``dog'' or ``cat'', for which correct predictions may stem from memorized pre-trained semantic priors rather than genuine understanding of the demonstrations. 
In contrast, by requiring models to identify personalized concepts such as a dog named ``John'', \oursbenchmark provides a more reliable test of whether MLLMs truly understand and leverage multimodal ICL demonstrations. 

Our main contributions are:
\vspace{-0.75em}
\begin{itemize}[leftmargin=12pt, itemsep=-0.5pt]
    \item We show that prior multimodal ICL methods suffer from limited generalization and forgetting.
    
    \item We identify weak visual grounding as a key bottleneck in multimodal ICL and propose \ours, a proposition-level contrastive reward that improves grounding without ICL-formatted data.
    
    
    \item We introduce \oursbenchmark, a personalized multimodal ICL captioning benchmark for evaluating grounded use of demonstrations.    
\end{itemize}
\vspace{-0.5em}


\section{Related Work}
\label{sec:related_work}

\textbf{Multimodal In-context Learning.}
Building on the emergence of ICL in LLMs, prior work has attempted to extend ICL to MLLMs, primarily by improving the demonstrations or by adapting the model with ICL-formatted data.
Retrieval- and ordering-based methods select demonstrations that are similar to the query~\cite{baldassini2024makes, luo2024does, chen2025can}, yet multimodal ICL remains highly sensitive to prompt design, sample order, and cross-modal similarity estimates~\cite{lu2022fantastically, qin2024factors, yang2024lever, guo2025deepseek}.
As a result, they often underperform random selection~\cite{li2025m, yang2024lever}.
ICL training-based methods instead expose MLLMs to ICL-formatted datasets, either through full-model tuning or additional learnable modules~\cite{doveh2024towards, tai2024link, chen2025true, li2025m, jiang2025mimic}.
These methods can improve performance on the same ICL datasets used for adaptation, but often generalize poorly to unseen ICL benchmarks and can forget previously well-performed tasks (Fig.~\ref{fig:forgetting}).

Task vector-based methods compress demonstrations into activation-space vectors that replicate the effect of steering hidden states during ICL, and apply them at inference in place of demonstrations to steer hidden states~\cite{peng2024live, liu2024context, todd2024function}.
They reduce context and memory overhead, but potentially losing fine-grained visual information needed for complex multimodal tasks~\cite{huang2024multimodal, ma2026and}.
Furthermore, similar to ICL training-based methods, they show limited generalization to unseen ICL benchmarks (Fig.~\ref{fig:teaser}). 
In contrast to these ICL data-dependent routes, \ours targets a more fundamental bottleneck, \ie, poor visual grounding, improving multimodal ICL without relying on ICL-formatted examples.

\textbf{Visual Grounding and Hallucination in MLLMs.}
MLLMs often over-rely on language priors and textual context, producing hallucinated objects, attributes, or relations not supported by the image~\cite{wang2023evaluation, liu2024mitigating, gunjal2024detecting, yu2024hallucidoctor}.
This is critical for multimodal ICL: if a model does not ground generated text in relevant visual evidence, it cannot reliably leverage the visual information in demonstrations~\cite{shukor2024beyond}.
Prior hallucination-mitigation methods include improving instruction-tuning data~\cite{liu2024mitigating, wang2024vigc, yu2024hallucidoctor} and adding auxiliary detectors or revisers~\cite{zhou2024analyzing, wang2024mitigating, heiman2025factchexcker}.
Despite their effectiveness, they often require access to large-scale instruction data and incur substantial parameter and computational overhead~\cite{zhu2025ibd}.

Alternatively, visual grounding approaches~\cite{liu2024paying, gao2025aim, chen2025true}, often directly amplify attention to visual tokens to counter the strong bias of MLLMs toward textual context over visual inputs.
However, directly scaling visual token norms may increase attention to irrelevant visual tokens~\cite{jung2025visual} and potentially distort feature representations by altering embedding magnitudes~\cite{zeng2021improving}.
To avoid this, rather than modifying token norms, \ours enhances visual grounding through a contrastive reward that promotes capturing discriminative visual cues from image–caption pairs, encouraging generated content to be supported by the corresponding image rather than by language priors alone.

See Sec.~\ref{sec:appx:related_work} for detailed related work on multimodal ICL and hallucination mitigation methods.

\section{Approach}
We address multimodal ICL from the perspective of visual grounding. 
Our key idea is that MLLMs fail at multimodal ICL not primarily because they lack ICL-specific supervision, but because they often fail to ground demonstrations in the relevant visual evidence. 
The weak grounding leads to over-reliance on textual cues and leads to hallucinations. 
To address this, we propose \ours, a simple yet effective reinforcement learning framework that improves visual grounding via contrastive rewards on visually contrastive image pairs. 
The key intuition is that coarse captions missing fine-grained visual details between two similar images (\eg, ``A hydrant on the road'' in Fig.~\ref{fig:overview}) can also match the wrong image, leading to lower rewards. 
Consequently, optimizing this objective encourages the model to capture discriminative visual details and ground generation in the image. 


However, rewarding only discriminative differences can bias the model toward shortcut descriptions that ignore visual details shared across the two images (\eg, ``on the sidewalk next to a sign'' in Fig.~\ref{fig:overview}), since describing shared content increases similarity to negative samples and reduces the reward. 
To overcome this, \ours further decomposes generated captions into propositions and applies the contrastive rewards only to image-specific propositions, while separately encouraging shared propositions to align with content common to both images. 
This proposition-level design allows the model to preserve complete visual descriptions while still learning to distinguish subtle differences. 
We provide an overview of \ours in Fig.~\ref{fig:overview} and a pseudocode in Sec.~\ref{sec:appx:pseoducode}.

\subsection{Multimodal Contrastive Training Pairs}
\begin{figure}[t!]
    \centering
    \includegraphics[width=\columnwidth]{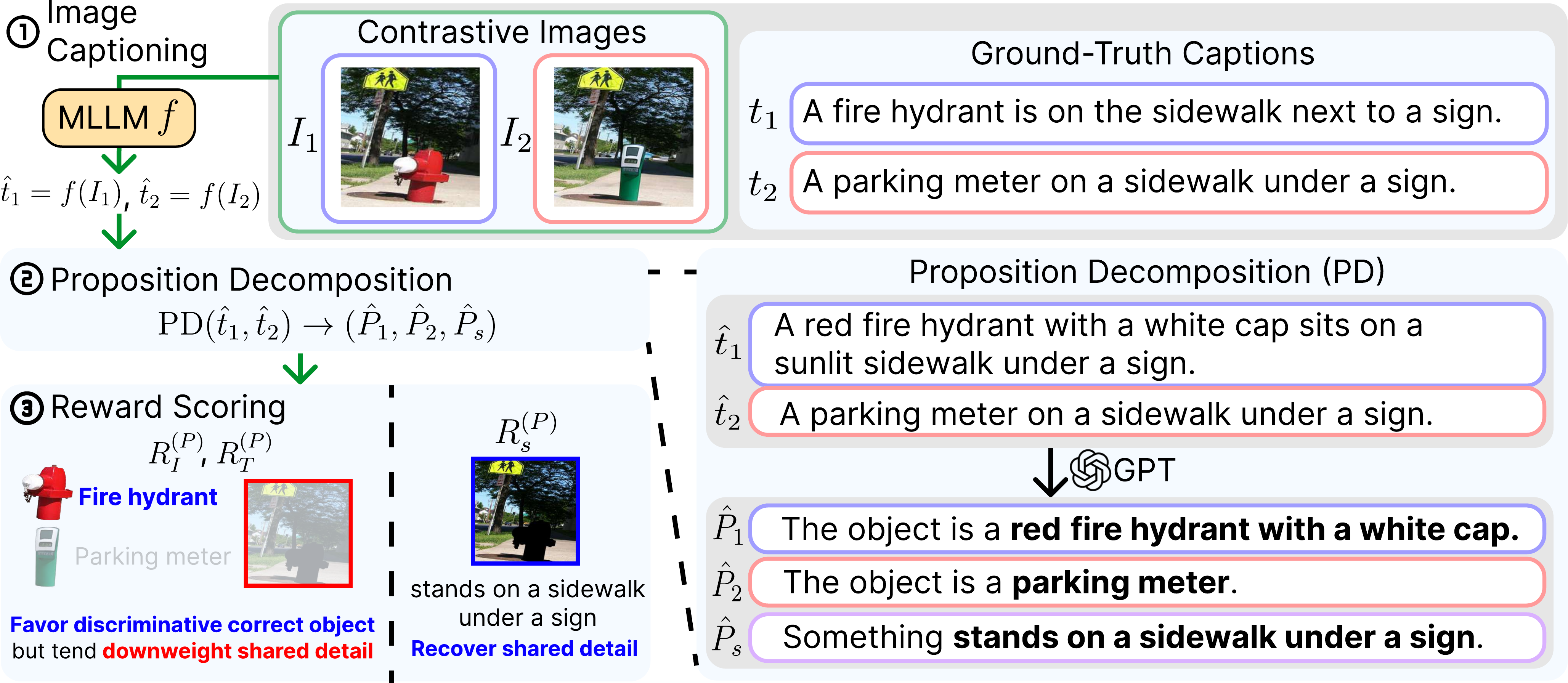}
    \vspace{-1em}
    \caption{
    \textbf{Overview of \ours.}
    (1) MLLM $f$ predict captions $\hat{t}_1, \hat{t}_2$ for contrastive images.
    (2) Decompose them into propositions using ChatGPT: unique to $I_1$ (\ie, $\hat{P}_1$), $I_2$ (\ie, $\hat{P}_2$), or shared (\ie, $\hat{P}_s$).
    (3) Compute proposition-level contrastive reward: $R_I$ and $R_T$ encourage $\hat{t}_1$ to reflect discriminative visual details (\ie, fire hydrant vs. parking meter), while $R_s$ preserves shared details.
    }
    \label{fig:overview}
\end{figure}

\ours is trained on visually contrastive image pairs that differ in fine-grained visual content such object identity or attributes, along with their captions, as illustrated in Fig.~\ref{fig:overview} (more in  Fig.~\ref{fig:contrastive_data_examples}). 
Formally, each sample consists of two image-text pairs, $(I_1, t_1)$ and $(I_2, t_2)$, where captions $t_1$ and $t_2$ describe images $I_1$ and $I_2$, respectively. 
Our method is compatible with any multimodal contrastive dataset containing such image-caption pairs. 
In this work, we use widely used multimodal contrastive datasets, InpaintCOCO~\cite{rosch2024enhancing} and EQBEN~\cite{wang2023equivariant}, with additional results with other datasets in Sec.~\ref{sec:appx:cont_datasets}.

\subsection{Contrastive Reward Function}
\textbf{Sentence-level Contrastive Reward.} 
We first introduce a sentence-level contrastive reward that encourages MLLMs to capture fine-grained differences between visually similar images. 
The intuition is that a caption grounded in the correct image should be more similar to its corresponding image and caption than to a non-corresponding one. 
By rewarding this contrast, the model is encouraged to attend to discriminative visual evidence rather than generate coarse descriptions. 
Formally, given a contrastive sample $\{(I_1, t_1), (I_2, t_2)\}$, an MLLM $f$ first generates captions $\hat{t}_1 = f(I_1)$ and $\hat{t}_2 = f(I_2)$. 
We then define an image-text contrastive reward as:
\begin{equation}
    \label{eq:reward_I}
    R_I(I_1, I_2, \hat{t}_1, \hat{t}_2) =
    \sigma\left(
    (\text{sim}_I(I_1, \hat{t}_1) - \text{sim}_I(I_2, \hat{t}_1))
    +
    (\text{sim}_I(I_2, \hat{t}_2) - \text{sim}_I(I_1, \hat{t}_2))
    \right),
\end{equation}
\vspace{-0.25em}
where $\sigma$ is the sigmoid function and $\text{sim}_I(I, t)=\cos(f_v(I), f_t(t))$ denotes cosine similarity between image and text embeddings produced by a vision encoder $f_v$ and a text encoder $f_t$. We use CLIP-ViT-B/32~\cite{radford2021learning} as $f_v$ and the CLIP text encoder as $f_t$. 
This reward encourages each generated caption to align with its corresponding image while discouraging alignment with the non-corresponding image.

In addition, we define a text-text contrastive reward that encourages the generated captions to match their corresponding ground-truth captions while remaining dissimilar to non-corresponding captions: 
\begin{equation}
    \label{eq:reward_T}
    R_T(t_1, t_2, \hat{t}_1, \hat{t}_2) =
    \sigma\left(
    (\text{sim}_T(t_1, \hat{t}_1) - \text{sim}_T(t_2, \hat{t}_1))
    +
    (\text{sim}_T(t_2, \hat{t}_2) - \text{sim}_T(t_1, \hat{t}_2))
    \right),
\end{equation}
where $\text{sim}_T(t_i,t_j)=\cos(f_t(t_i),f_t(t_j))$ is cosine similarity between text embeddings.

The final sentence-level reward is $R = \lambda_I R_I + \lambda_T R_T$ where $\lambda_I$ and $\lambda_T$ control the relative importance of the two terms. 
We optimize this reward using Group Relative Policy Optimization (GRPO)~\cite{shao2024deepseekmath}.

\paragraph{\oursfull.}
\label{subsec:ours_cont_reward}

Sentence-level reward encourages discriminative grounding, but it can also bias the model toward incomplete descriptions. 
Since shared content appears in both images, describing it may increase similarity to the non-corresponding sample and thus reduce the reward. 
For example, in Fig.~\ref{fig:overview}, the images differ only in the object (\ie, hydrant versus parking meter), while other details such as the sidewalk and nearby sign are shared. 
Describing this shared scene content (\eg, ``on the sidewalk next to a sign'') increases similarity to both images and may therefore lower the reward. 
As a result, the model may learn shortcut captions that mention only the distinguishing difference (\eg, ``The object is a hydrant'') while omitting shared visual details.

To address this limitation, \ours computes contrastive rewards at the proposition level rather than on full-sentence captions. 
Specifically, we decompose the generated captions $\hat{t}_1$ and $\hat{t}_2$ into three types of propositions: (i) $\hat{P}_1$, content unique to $\hat{t}_1$; (ii) $\hat{P}_2$, content unique to $\hat{t}_2$; and (iii) $\hat{P}_s$, content shared by both captions (see Fig.~\ref{fig:overview} for an example), using GPT-4o-mini~\cite{Achiam2023GPT4TR}.
See Sec.~\ref{sec:appx:prompt_decomposition} for the details of decomposition process.
We then compute the proposition-level contrastive reward using only the image-specific propositions:
\begin{equation}
    \label{eq:new_reward_I}
    R^{(P)}_I(I_1, I_2, \hat{P}_1, \hat{P}_2) =
    \sigma\left(
    (\text{sim}_I(I_1, \hat{P}_1) - \text{sim}_I(I_1, \hat{P}_2))
    +
    (\text{sim}_I(I_2, \hat{P}_2) - \text{sim}_I(I_2, \hat{P}_1))
    \right).
\end{equation}
By excluding shared propositions from the contrastive terms, this formulation encourages the model to distinguish the two images without penalizing content that is true for both.

Similarly, we replace sentence-level text similarity with proposition-level entailment. 
Since propositions are sub-sentential units, cosine similarity between sentence embeddings is not ideal. Instead, we use an NLI-based entailment score $\text{entail}(t,\hat{P})$~\cite{Bowman2015ALA}, which measures how strongly the ground-truth caption $t$ entails the predicted proposition $\hat{P}$. 
The proposition-level text reward is then defined as:
\begin{equation}
    R^{(P)}_T(t_1, t_2, \hat{P}_1, \hat{P}_2) =\!
    \sigma\left(
    (\text{entail}(t_1, \hat{P}_1)\!- \!\text{entail}(t_2, \hat{P}_1))
    \!+\!
    (\text{entail}(t_2, \hat{P}_2) \!- \!\text{entail}(t_1, \hat{P}_2))
    \right).
\end{equation}

To preserve shared visual details, we further introduce a shared-content reward for $\hat{P}_s$:
\begin{equation}
    \label{eq:new_reward_T}
    R^{(P)}_s(I_1, I_2, t_1, t_2, \hat{P}_s) =
    \sigma\left(
    \text{sim}_I(I_1, \hat{P}_s) + \text{sim}_I(I_2, \hat{P}_s)
    +
    \text{entail}(t_1, \hat{P}_s) + \text{entail}(t_2, \hat{P}_s)
    \right).
\end{equation}
This term encourages $\hat{P}_s$ to capture content supported by both images and both ground-truth captions.

The final proposition-level reward used in \ours is below:
\begin{equation}
    \label{eq:objective}
    R^{(P)} = \lambda_I R^{(P)}_I + \lambda_T R^{(P)}_T + \lambda_s R^{(P)}_s,
\end{equation}
where $\lambda_I$, $\lambda_T$, and $\lambda_s$ are weighting coefficients.

\section{\oursbenchmark Benchmark}
\label{sec:benchmark}

\begin{figure}[t!]
    \centering
    \vspace{-1em}
    \includegraphics[width=\columnwidth]{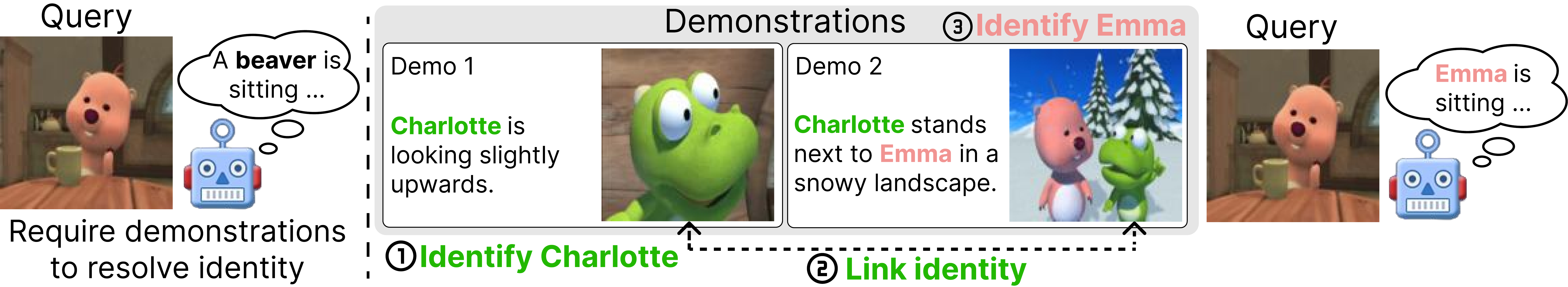}
    \vspace{-1.4em}
    \caption{
    \textbf{Example of 2-shot \oursbenchmark.} 
    (left) Query image alone cannot infer the personalized identity.
    (right) (1) Identifying the dinosaur as ``Charlotte'' in the first demo and (2) linking it to the dinosaur in the second enables (3) inferring that the pink beaver is ``Emma'' in the second, allowing correct prediction of the personalized identifier for the query.
    See Fig.~\ref{fig:additional_examples_picl} and Fig.~\ref{fig:picl_shot} for more examples.
    }
    \label{fig:ours_benchmark}
    \vspace{-0.5em}
\end{figure}

Standard multimodal ICL benchmarks~\cite{chen2025true, lin2014microsoft, zhao2024mmicl, zong2025vlicl} often involve generic identifiers (\eg, `dog', `he') that MLLMs have likely seen during pre-training.
As a result, models can answer correctly using pre-trained priors without truly leveraging demonstrations.
\oursbenchmark mitigates this shortcut by evaluating personalized multimodal ICL captioning, where demonstrations introduce identifiers not recoverable from the query image alone (\eg, particular character names).
Correct prediction requires grounding these personalized identifiers in the demonstration images and transfer that association to the query.

We construct \oursbenchmark from visual storytelling datasets—PororoSV~\cite{li2019storygan}, FlintstonesSV~\cite{gupta2018imagine}, and StoryStream~\cite{yang2025seed}—where the same characters appear across frames, each with  caption.
We first manually select frames to form demonstrations and queries.
We then replace original identifiers in the captions with personalized identifiers, randomly sampled via GPT-4o-mini~\cite{Achiam2023GPT4TR}, requiring models to use the correct identifier when the same entity appears in the query.
This benchmark thus complements standard ICL evaluations by testing whether models truly leverage visual context, rather than merely recognizing familiar categories.
From each dataset (PororoSV, FlintstonesSV, and StoryStream), we form 1,000 training samples (for ICL data-dependent methods) and 100 test samples, yielding 3,300 samples per shot setting (1-4 shot).
Notably, higher-shot settings are designed to require reasoning over increasingly complex relationships among demonstrations (\eg, Fig.~\ref{fig:picl_shot}), rather than simply providing more context; thus, performance does not necessarily improve with more shots.
See Fig.~\ref{fig:ours_benchmark} for an example (more in Sec.~\ref{sec:appx:benchmark_examples}) and Sec.~\ref{sec:appx:our_benchmark_details} for more details.




\section{Experiments}
\subsection{Experimental Setups}
\label{sec:subsec:main_exp_setup}
\textbf{Models.}
We use Qwen-VL-3-8B~\cite{bai2025qwen3} as the main MLLM for all primary comparisons.
We further evaluate \ours across multiple MLLMs, including LLaVA-OneVision-1.5-8B~\cite{an2025llava} and InternVL-3.5-8B~\cite{wang2025internvl3} in Sec.~\ref{sec:appx:diverse_models}, as well as across diverse model sizes, including 4B and 14B (Sec.~\ref{sec:appx:diverse_sizes}).

\begin{table}[t!]
\centering
\small
\resizebox{\linewidth}{!}{
\begin{tabular}{lclcccccccccc}
\toprule
\multirow{2}{*}{\textbf{Category}} & \multirow{2}{*}{\makecell{\textbf{ICL Training}\\ \textbf{Dataset}}} & \multirow{2}{*}{\textbf{Method}} & \multicolumn{5}{c}{\textbf{MS COCO}} & \multicolumn{5}{c}{\textbf{PICL}} \\ 
& & & 0-shot & 1-shot & 2-shot & 4-shot & 8-shot & 0-shot & 1-shot & 2-shot & 3-shot & 4-shot \\

\cmidrule(lr){1-1} \cmidrule(lr){2-2} \cmidrule(lr){3-3} \cmidrule(lr){4-8} \cmidrule(lr){9-13}

Off-the-shelf & \multirow{1}{*}{-} & - 
& 15.53 & 19.02 & 21.21 & 23.52 & 25.65
& 24.50 & 25.40 & 24.78 & 28.30 & 29.29 \\

\cmidrule(lr){1-1} \cmidrule(lr){2-2} \cmidrule(lr){3-3} \cmidrule(lr){4-8} \cmidrule(lr){9-13}

\multirow{3}{*}{\makecell{Retrieval and \\ ordering}} & \multirow{3}{*}{-}
& RICES {\footnotesize \color{blue}{(NeurIPS22)}}
& 15.53 & 19.02 & 21.08 & 23.36 & 25.70
& 24.50 & 25.40 & 26.44 & 27.14 & 28.20 \\

& & VICL {\footnotesize \color{blue}{(ACL24)}}
& 17.41 & 21.52 & 23.44 & 24.16 & 27.63
& 23.12 & 26.77 & 27.38 & 25.96 & 29.41 \\

& & MMICES {\footnotesize \color{blue}{(WACV25)}}
& 15.53 & 19.02 & 21.65 & 22.52 & 25.16
& 24.50 & 25.40 & 21.70 & 23.82 & 24.79 \\

\cmidrule(lr){1-1} \cmidrule(lr){2-2} \cmidrule(lr){3-3} \cmidrule(lr){4-8} \cmidrule(lr){9-13}

\multirow{4}{*}{Task vector} & \multirow{2}{*}{MS COCO}
& MTV {\footnotesize \color{blue}{(NeurIPS24)}}
& 15.84 & 18.36 & 18.68 & 18.53 & 19.02
& 12.11 & 15.02 & 16.41 & 15.88 & 17.63 \\

& & FV {\footnotesize \color{blue}{(ICLR24)}}
& 14.56 & 17.97 & 20.65 & 22.93 & 25.33
& 22.31 & 23.88 & 24.12 & 26.72 & 28.41 \\

\cmidrule(lr){2-2} \cmidrule(lr){3-3} \cmidrule(lr){4-8} \cmidrule(lr){9-13}

& \multirow{2}{*}{PICL}
& MTV {\footnotesize \color{blue}{(NeurIPS24)}}
& 9.11 & 14.03 & 13.74 & 15.88 & 12.84
& 9.18 & 14.87 & 13.96 & 15.27 & 13.02 \\

& & FV {\footnotesize \color{blue}{(ICLR24)}}
& 10.73 & 13.88 & 16.21 & 18.44 & 20.91 
& 24.04 & 25.11 & 25.57 & 28.50 & 29.84 \\

\cmidrule(lr){1-1} \cmidrule(lr){2-2} \cmidrule(lr){3-3} \cmidrule(lr){4-8} \cmidrule(lr){9-13}

\multirow{8}{*}{ICL training} & \multirow{4}{*}{MS COCO}
& MT-ICL {\footnotesize \color{blue}{(ECCVW24)}}
& 37.70 & 42.95 & 42.76 & 42.37 & 42.48
& 30.41 & 29.72 & 30.88 & 31.26 & 32.02 \\

& & DARA {\footnotesize \color{blue}{(COLM25)}}
& 25.27 & 29.10 & 30.83 & 32.25 & 38.40
& 22.03 & 23.84 & 23.16 & 25.72 & 26.19 \\

& & MimIC {\footnotesize \color{blue}{(CVPR25)}}
& 29.38 & 36.06 & 35.64 & 35.51 & 37.21
& 21.88 & 26.41 & 24.92 & 26.76 & 29.03 \\

& & STV {\footnotesize \color{blue}{(AAAI26)}}
& \underline{38.72} & \underline{44.42} & \underline{42.33} & \underline{43.04} & \underline{42.95}
& 29.43 & 30.01 & 30.88 & 31.92 & 32.27 \\

\cmidrule(lr){2-2} \cmidrule(lr){3-3} \cmidrule(lr){4-8} \cmidrule(lr){9-13}

& \multirow{4}{*}{PICL}
& MT-ICL {\footnotesize \color{blue}{(ECCVW24)}}
& 34.18 & 39.77 & 40.12 & 39.02 & 39.88
& \underline{32.63} & 32.63 & 32.59 & 33.73 & 34.55 \\

& & DARA {\footnotesize \color{blue}{(COLM25)}}
& 21.92 & 25.84 & 27.66 & 29.11 & 35.07
& 24.50 & 25.22 & 25.47 & 28.20 & 29.14 \\

& & MimIC {\footnotesize \color{blue}{(CVPR25)}}
& 25.74 & 32.88 & 31.92 & 31.73 & 33.44
& 24.56 & 28.53 & 27.71 & 28.85 & 31.41 \\

& & STV {\footnotesize \color{blue}{(AAAI26)}}
& 35.41 & 41.26 & 39.74 & 40.12 & 40.33
& 32.11 & \underline{32.99} & \underline{32.61} & \underline{34.48} & \underline{34.74} \\

\cmidrule(lr){1-1} \cmidrule(lr){2-2} \cmidrule(lr){3-3} \cmidrule(lr){4-8} \cmidrule(lr){9-13}

\multirow{7}{*}{\makecell{Mitigating \\ hallucination}} & \multirow{7}{*}{-}
& PAI  {\footnotesize \color{blue}{(ECCV24)}}
& 13.92 & 20.11 & 18.84 & 24.33 & 23.07
& 21.74 & 24.62 & 23.18 & 25.11 & 27.89 \\

& & VOLCANO  {\footnotesize \color{blue}{(NAACL24)}} 
& 31.72 & 38.41 & 33.88 & 37.12 & 41.04
& 23.47 & 26.82 & 24.91 & 27.66 & 29.84 \\

& & LURE {\footnotesize \color{blue}{(ICLR24)}} 
& 16.41 & 17.36 & 19.58 & 22.07 & 24.12
& 22.18 & 27.04 & 25.32 & 29.88 & 30.17 \\

& & LRV {\footnotesize \color{blue}{(ICLR24)}}
& 17.41 & 29.22 & 26.77 & 23.89 & 30.26
& 26.73 & 31.21 & 28.96 & 30.41 & 32.55 \\

& & IBD  {\footnotesize \color{blue}{(CVPRW25)}} 
& 30.94 & 34.28 & 37.51 & 38.67 & 35.88
& 20.33 & 23.91 & 22.48 & 23.76 & 26.12 \\

& & SPARC {\footnotesize \color{blue}{(ICML25)}} 
& 33.18 & 37.73 & 34.26 & 39.84 & 42.11
& 24.91 & 30.12 & 26.44 & 27.03 & 33.08 \\

& & \ours (Ours)
& \textbf{39.12} & \textbf{46.88} & \textbf{46.41} & \textbf{46.26} & \textbf{46.03}
& \textbf{33.54} & \textbf{36.11} & \textbf{35.89} & \textbf{36.71} & \textbf{36.76} \\

\bottomrule
\end{tabular}
}
\vspace{0.5em}
\caption{
\textbf{Comparison in multimodal ICL captioning benchmarks.}
Best and second-best scores are bolded and underlined per dataset and shot.
We use Qwen-VL-3-8B and apply random ordering of demonstrations for all baselines except retrieval- and ordering-based methods.
}
\vspace{-.5em}
\label{tab:icl_captioning_qwen}
\end{table}

\textbf{Baselines.}
We compare \ours with SOTA multimodal ICL methods: (\romannumeral 1) retrieval- and ordering-based methods (RICES~\cite{alayrac2022flamingo}, VICL~\cite{zhou2024visual}, MMICES~\cite{chen2025can}), (\romannumeral 2) ICL training-based methods (STV~\cite{ma2026and}, MimIC~\cite{jiang2025mimic}, MT-ICL~\cite{doveh2024towards}, DARA~\cite{chen2025true}), and (\romannumeral 3) task vector-based methods (MTV~\cite{huang2024multimodal}, FV~\cite{todd2024function}).
We also compare with hallucination-mitigation baselines (PAI~\cite{liu2024paying}, VOLCANO~\cite{lee2024volcano}, LURE~\cite{zhou2023analyzing}, LRV~\cite{liu2024mitigating}, IBD~\cite{zhu2025ibd}, SPARC~\cite{jung2025visual}) to separate gains from improved visual grounding from those of generic hallucination reduction.
For fair comparison, all training-based baselines (including \ours) use the same number of training samples, though ICL data-dependent methods use ICL datasets while others use non-ICL data, as more detailed in Sec.~\ref{sec:appx:exp_details}.
The off-the-shelf model with random demonstrations is included as a reference. 
Further baseline details are in Sec.~\ref{sec:appx:details_baselines}.


\textbf{Metrics.}
We report accuracy for multimodal ICL classification and ROUGE-L~\cite{lin-2004-rouge} for multimodal ICL captioning, as a commonly used metric for evaluating generation quality against ground truth~\cite{yang2024dualpersonalizing, seo2026colora}.
Moreover, to quantify visual grounding, we use target attention score (TAS)~\cite{ou2026mllms}, which averages attention from each generated token to the corresponding target-region image tokens over attention heads; higher TAS indicates stronger region-token grounding.
Target regions are obtained from SAM3 segmentation maps~\cite{carion2025sam}.
For hallucination analysis, we report OpenCHAIR scores, where lower values indicate fewer object hallucinations.
See Sec.~\ref{sec:appx:exp_details} for more details of metrics.


\textbf{Benchmarks.}
For multimodal ICL classification benchmarks, we evaluate on TrueMICL~\cite{chen2025can} and ISEKAI~\cite{tai2024link}. 
For multimodal ICL captioning, we employ MS COCO~\cite{lin2014microsoft} and our proposed personalized ICL benchmark, \oursbenchmark, where demonstrations introduce personalized character names that must be visually grounded.
We also evaluate on MIC~\cite{zhao2024mmicl} and VL-ICL~\cite{zong2025vlicl} in Sec.~\ref{sec:appx:additional_ICL_classification_results}.

\subsection{Quantitative Analysis}

\textbf{Multimodal ICL Captioning Tasks.}
Table~\ref{tab:icl_captioning_qwen} evaluates whether stronger visual grounding improves ICL captioning.
As shown, \ours significantly outperforms baselines across MS COCO and \oursbenchmark at all settings, despite using no ICL-formatted training data.
These results support our hypothesis: strong visual grounding elicits stronger ICL capability without directly fitting to specific ICL benchmarks.
In contrast, ICL data-dependent methods perform well on the benchmark used for training or task-vector extraction,  but degrades significantly on other ICL datasets, often falling below the off-the-shelf model, indicating limited generalization.
Retrieval- and ordering-based methods often underperform the random selection (\ie, off-the-shelf model), as also observed in prior work~\cite{yang2024lever, li2025m}.


Moreover, hallucination-mitigation baselines show only marginal gains and often underperform the off-the-shelf model. 
We attribute this to their approach of enhancing visual grounding through direct amplification of visual attention norms, which fails to focus on task-relevant regions (Sec.~\ref{sec:appx:attention_quali}).
In contrast, \ours naturally enhances visual grounding through a contrastive reward, resulting in more accurate visual grounding.
See Sec.~\ref{sec:appx:additional_ICL_captioning_results} for additional ICL captioning additional ICL captioning results and analysis of personalized identifier accuracy.


\begin{table}[t!]
\centering
\small
\resizebox{\linewidth}{!}{
\begin{tabular}{lclcccccccccc}
\toprule
\multirow{2}{*}{\textbf{Category}} & \multirow{2}{*}{\makecell{\textbf{ICL Training}\\ \textbf{Dataset}}} & \multirow{2}{*}{\textbf{Method}} & \multicolumn{5}{c}{\textbf{TrueMICL}} & \multicolumn{5}{c}{\textbf{ISEKAI}} \\ 
& & & 0-shot & 1-shot & 2-shot & 4-shot & 8-shot & 0-shot & 2-shot & 4-shot & 8-shot & 12-shot \\

\cmidrule(lr){1-1} \cmidrule(lr){2-2} \cmidrule(lr){3-3} \cmidrule(lr){4-8} \cmidrule(lr){9-13}

Off-the-shelf & \multirow{1}{*}{-} & - 
& 27.11 & 44.79 & 48.83 & 55.03 & 62.70
& 0.16 & 37.98 & 83.68 & 93.79 & 96.42 \\

\cmidrule(lr){1-1} \cmidrule(lr){2-2} \cmidrule(lr){3-3} \cmidrule(lr){4-8} \cmidrule(lr){9-13}

\multirow{3}{*}{\makecell{Retrieval and \\ ordering}} & \multirow{3}{*}{-}
& RICES {\footnotesize \color{blue}{(NeurIPS22)}}
& 27.11 & 44.79 & 47.16 & 51.83 & 61.86
& 0.16 & 36.31 & 81.21 & 90.53 & 92.36 \\

& & VICL {\footnotesize \color{blue}{(ACL24)}}
& 28.22 & 45.44 & 50.01 & 55.12 & 61.80
& 1.12 & 37.81 & 74.64 & 88.93 & 88.41 \\

& & MMICES {\footnotesize \color{blue}{(WACV25)}}
& 27.11 & 44.79 & 48.40 & 58.07 & 63.04
& 0.16 & 34.55 & 73.09 & 85.27 & 84.87 \\

\cmidrule(lr){1-1} \cmidrule(lr){2-2} \cmidrule(lr){3-3} \cmidrule(lr){4-8} \cmidrule(lr){9-13}

\multirow{4}{*}{Task vector} & \multirow{2}{*}{TrueMICL}
& MTV {\footnotesize \color{blue}{(NeurIPS24)}}
& 23.80 & 36.45 & 42.52 & 45.30 & 55.70
& 3.04 & 32.88 & 81.98 & 94.99 & 90.12 \\

& & FV {\footnotesize \color{blue}{(ICLR24)}}
& 28.82 & 43.88 & 48.61 & 55.87 & 60.65
& 0.96 & 31.58 & 78.78 & 93.19 & \underline{97.92} \\

\cmidrule(lr){2-2} \cmidrule(lr){3-3} \cmidrule(lr){4-8} \cmidrule(lr){9-13}

& \multirow{2}{*}{ISEKAI}
& MTV {\footnotesize \color{blue}{(NeurIPS24)}}
& 21.82 & 40.11 & 44.27 & 48.39 & 54.92
& 2.15 & 28.65 & 84.55 & 94.11 & 96.58 \\

& & FV {\footnotesize \color{blue}{(ICLR24)}}
& 26.44 & 38.76 & 41.98 & 52.02 & 55.18
& 1.21 & 35.44 & 84.72 & 92.18 & 97.03 \\

\cmidrule(lr){1-1} \cmidrule(lr){2-2} \cmidrule(lr){3-3} \cmidrule(lr){4-8} \cmidrule(lr){9-13}

\multirow{8}{*}{ICL training} & \multirow{4}{*}{TrueMICL}
& MT-ICL {\footnotesize \color{blue}{(ECCVW24)}}
& 28.12 & 45.26 & 49.96 & \underline{57.80} & 62.59
& 3.47 & 31.72 & 70.94 & 93.18 & 84.02 \\

& & DARA {\footnotesize \color{blue}{(COLM25)}}
& \underline{29.11} & 45.99 & 50.86 & 56.20 & \underline{65.93}
& 0.82 & 27.64 & 79.11 & 84.27 & 91.08 \\

& & MimIC {\footnotesize \color{blue}{(CVPR25)}}
& 27.08 & \underline{48.62} & \underline{54.65} & 55.00 & 61.21
& 3.02 & 34.86 & 81.03 & 85.51 & 92.19 \\

& & STV {\footnotesize \color{blue}{(AAAI26)}}
& 29.17 & 45.35 & 49.78 & 57.62 & 64.33
& \textbf{5.73} & 24.39 & 72.58 & 90.92 & 86.44 \\

\cmidrule(lr){2-2} \cmidrule(lr){3-3} \cmidrule(lr){4-8} \cmidrule(lr){9-13}

& \multirow{4}{*}{ISEKAI}
& MT-ICL {\footnotesize \color{blue}{(ECCVW24)}}
& 21.94 & 38.21 & 40.47 & 49.89 & 58.36
& 0.40 & 48.06 & 78.14 & \underline{96.46} & 95.58 \\

& & DARA {\footnotesize \color{blue}{(COLM25)}}
& 24.02 & 39.15 & 44.06 & 48.58 & 55.84
& 1.51 & 40.41 & 75.92 & 87.63 & 88.77 \\

& & MimIC {\footnotesize \color{blue}{(CVPR25)}}
& 27.08 & 38.74 & 43.55 & 51.12 & 56.76
& 0.16 & 44.87 & 83.02 & 94.74 & 94.11 \\

& & STV {\footnotesize \color{blue}{(AAAI26)}}
& 19.82 & 37.41 & 40.02 & 47.73 & 53.26
& 2.92 & \underline{49.83} & \underline{86.02} & 96.11 & 95.77 \\

\cmidrule(lr){1-1} \cmidrule(lr){2-2} \cmidrule(lr){3-3} \cmidrule(lr){4-8} \cmidrule(lr){9-13}

\multirow{7}{*}{\makecell{Mitigating \\ hallucination}} & \multirow{7}{*}{-}
& PAI  {\footnotesize \color{blue}{(ECCV24)}}
& 24.91 & 44.12 & 47.83 & 56.02 & 59.67
& 1.87 & 39.22 & 73.11 & 91.54 & 90.02 \\

& & VOLCANO  {\footnotesize \color{blue}{(NAACL24)}} 
& 31.41 & 47.76 & 50.68 & 57.43 & 64.12
& 0.21 & 42.08 & 74.96 & 94.73 & 92.11 \\

& & LURE {\footnotesize \color{blue}{(ICLR24)}} 
& 24.08 & 42.31 & 45.92 & 52.77 & 60.18 
& 0.16 & 33.74 & 80.21 & 89.46 & 92.87 \\

& & LRV {\footnotesize \color{blue}{(ICLR24)}}
& 29.02 & 45.11 & 51.94 & 57.27 & 63.46
& 0.40 & 39.77 & 78.11 & \textbf{97.42} & 95.06 \\

& & IBD  {\footnotesize \color{blue}{(CVPRW25)}} 
& 27.21 & 45.27 & 50.02 & 55.44 & 62.18
& 1.42 & 45.63 & 72.88 & 91.56 & 89.47 \\

& & SPARC {\footnotesize \color{blue}{(ICML25)}} 
& 25.38 & 46.21 & 47.09 & 56.82 & 60.94
& 0.16 & 45.63 & 83.02 & 92.11 & 93.63 \\

& & \ours (\textbf{Ours})
& \textbf{33.02} & \textbf{51.47} & \textbf{56.11} & \textbf{62.26} & \textbf{68.08}
& \underline{3.94} & \textbf{50.72} & \textbf{86.23} & 96.18 & \textbf{98.41} \\

\bottomrule
\end{tabular}
}
\vspace{0.5em}
\caption{
\textbf{Comparison in multimodal ICL classification benchmarks.}
Best and second-best scores are bolded and underlined per dataset and shot.
We use Qwen-VL-3-8B and apply random ordering of demonstrations for all baselines except retrieval- and ordering-based methods.
}
\vspace{-1.25em}
\label{tab:icl_classification_qwen}
\end{table}

\begin{figure}[t!]
    \vspace{-0.25em}
    \centering   
    \includegraphics[width=\linewidth]{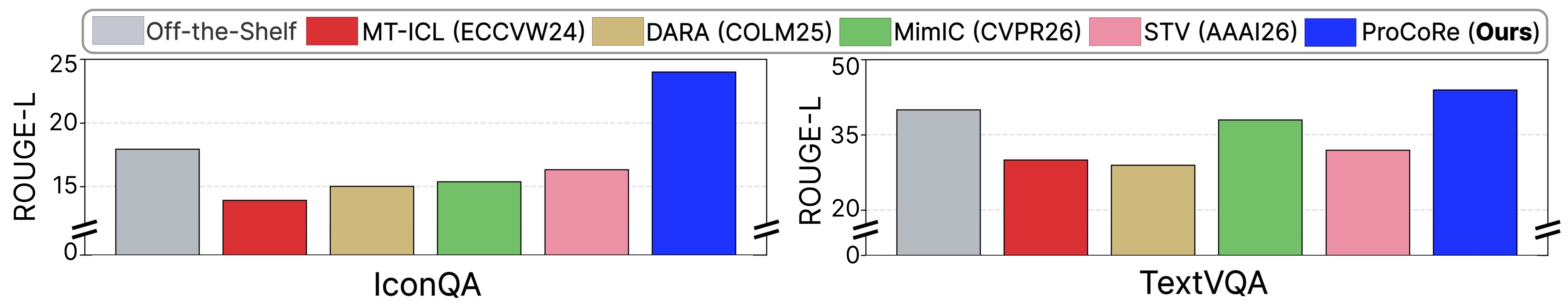}
    \vspace{-1.7em}
    \caption{
        \textbf{Effect of ICL training on simple VQA tasks.}
        ICL training-based methods, which perform additional instruction tuning on ICL datasets, degrade performance on previously well-solved simple VQA tasks (\ie, IconQA, TextVQA) due to forgetting.
        In contrast, \ours improves these tasks by enhancing visual grounding, while also achieving strong multimodal ICL performance (Tab.~\ref{tab:icl_classification_qwen}).
        }
    \label{fig:forgetting}
    \vspace{-0.5em}
\end{figure}

\begin{figure}[t!]
    \vspace{-0.2em}
    \centering   
    \includegraphics[width=\linewidth]{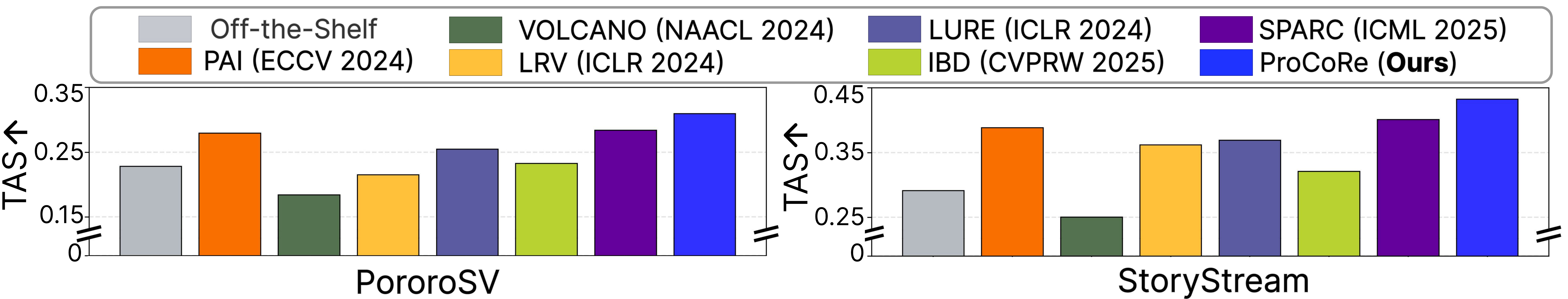}
    \vspace{-1.5em}
    \caption{
        \textbf{Comparison of target attention score (TAS) on multimodal ICL captioning benchmarks.}
        While hallucination-mitigation baselines sometimes enhance TAS compared to off-the-shelf model, our proposed \ours significantly outperforms both off-the-shelf model and baselines. 
        }
    \label{fig:TAS}
    \vspace{-0.5em}
\end{figure}

\textbf{Multimodal ICL Classification Tasks.}
We also evaluate \ours on ICL classification tasks and report results in Tab.~\ref{tab:icl_classification_qwen}. 
As shown in the table, consistent with the ICL captioning task, \ours significantly outperforms baselines across all shot settings, demonstrating that \ours generalizes strongly across diverse ICL tasks, while ICL data-dependent baselines continue to show limited generalization.
We provide additional ICL classification results in Sec.~\ref{sec:appx:additional_ICL_classification_results}.

\textbf{Visual Grounding and Hallucination Evaluation.}
Next we test whether the ICL gains in \ours come with strong visual grounding.
Visual attention map (Fig.~\ref{fig:highlevel-teaser} and Fig.~\ref{fig:quali_vg}) shows that \ours attends to relevant visual regions when generating response tokens.
Moreover, Fig.~\ref{fig:TAS} shows that \ours achieves the highest TAS, indicating stronger focus on task-relevant regions.

\begin{figure}[t!]
    \centering   
    \includegraphics[width=\linewidth]{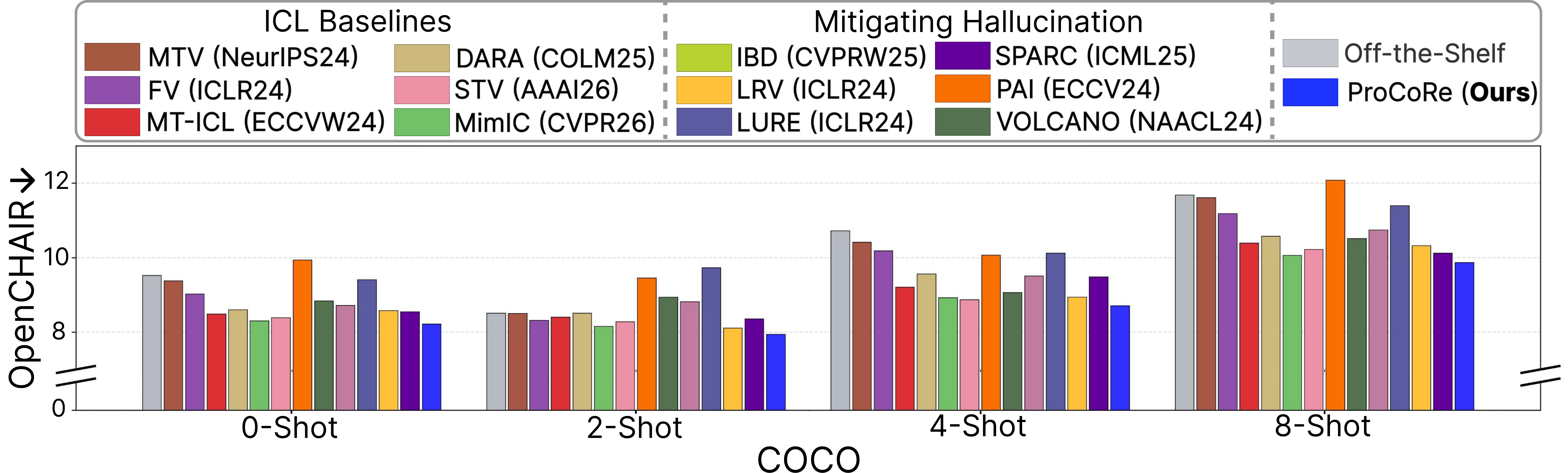}
    \vspace{-1.25em}
    \caption{
        \textbf{Comparison of hallucinations on MS COCO ICL captioning benchmark.}
        \ours consistently reduces hallucinations across varying numbers of demonstrations compared to baselines.
        }
    \label{fig:hallucination}
    \vspace{-0.5em}
\end{figure}

This improved visual grounding mitigates hallucination in multimodal ICL. 
As shown in Fig.~\ref{fig:hallucination}, \ours consistently shows lower OpenCHAIR scores than baselines. 
We believe that this reduction in hallucination, driven by strong visual grounding, contributes to strong multimodal ICL performance.
We provide further visual grounding analysis in Sec.~\ref{sec:appx:attention_quali}.

\textbf{General VQA Evaluation.}
We test whether general vision–language ability is preserved.
Figure~\ref{fig:forgetting} shows that \ours improves standard VQA performances, whereas ICL-training baselines often degrade.
This suggests that \ours does not trade general vision-language ability for benchmark-specific ICL behavior, but instead improves the fundamental visual grounding ability of MLLMs.

\textbf{Analysis across Diverse MLLMs and Model Sizes.}
In addition to Qwen-VL-3~\cite{bai2025qwen3}, we compare \ours with baselines across multiple MLLMs, including LLaVA-OneVision-1.5~\cite{an2025llava} and InternVL-3.5~\cite{wang2025internvl3} in Sec.~\ref{sec:appx:diverse_models}, as well as across diverse model sizes (4B, and 13B) in Sec.~\ref{sec:appx:diverse_sizes}, showing that \ours consistently achieves strong multimodal ICL performance across models and scales.

\textbf{Additional Experiments and Analysis.}
Additional results include: \ours across diverse contrastive benchmarks (Sec.~\ref{sec:appx:cont_datasets}) and hyperparameter analysis (Sec.~\ref{sec:appx:hyper_analysis}).

\textbf{Ablation Study.}
We analyze the contribution of \ours by comparing to (i) supervised fine-tuning (SFT) on contrastive pairs to disentangle gains beyond those achievable from contrastive-pair training alone, (ii) sentence-level contrastive rewards, to assess the effect of proposition-level decomposition in \ours.
As shown in Tab.~\ref{tab:ablation}, SFT with contrastive pairs shows only marginal gains, indicating that improvements in \ours stem from the proposed contrastive reward rather than contrastive-pair training itself.
Moreover, the significant gap between \ours and sentence-level rewards highlights the effectiveness of proposition-level contrastive rewards.



\begin{table*}[t]
\centering
\resizebox{\linewidth}{!}{
\begin{tabular}{lcccccccccc}
\toprule
\multirow{2}{*}{\textbf{Method}} 
& \multicolumn{5}{c}{\textbf{MS COCO}} 
& \multicolumn{5}{c}{\textbf{\oursbenchmark}} \\

& 0-shot & 1-shot & 2-shot & 4-shot & 8-shot
& 0-shot & 1-shot & 2-shot & 3-shot & 4-shot \\

\cmidrule(lr){1-1}
\cmidrule(lr){2-6}
\cmidrule(lr){7-11}

Off-the-Shelf (Qwen-VL-3-8B)
& 15.53 & 19.02 & 21.21 & 23.52 & 25.65
& 24.50 & 25.40 & 24.78 & 28.30 & 29.29 \\

SFT
& 19.48 & 27.92 & 30.11 & 26.84 & 37.62
& 22.13 & 27.88 & 23.41 & 30.74 & 26.81 \\

Sentence-Level Contrastive Reward
& 36.02 & 45.73 & 43.21 & 44.11 & 42.58
& 32.88 & 34.74 & 31.92 & 34.02 & 33.41 \\

\ours (\textbf{Ours})
& \textbf{39.12} & \textbf{46.88} & \textbf{46.41} & \textbf{46.26} & \textbf{46.03}
& \textbf{33.54} & \textbf{36.11} & \textbf{34.89} & \textbf{36.71} & \textbf{36.76} \\

\bottomrule
\end{tabular}
}
\vspace{-0.3em}
\caption{
    \textbf{Ablations of \ours.} 
    Gains in \ours stem from proposition-level contrastive rewards, not contrastive-pair training it self, with clear improvements over sentence-level rewards.
    }
\label{tab:ablation}
\vspace{-.6em}
\end{table*}

\section{Conclusions}
We show that existing multimodal ICL data-dependent methods suffer from limited generalization and catastrophic forgetting. 
Rather than relying on ICL benchmarks, we address a more fundamental issue: weak visual grounding in MLLMs.
To this end, we propose \ours, a grounding-first approach that strengthens visual–textual associations without using ICL-formatted data. 
By learning from visually contrastive pairs with proposition-level rewards, \ours promotes fine-grained grounding without shortcut descriptions.
Across ICL benchmarks, \ours improves performance and generalizes beyond the training data of ICL data-dependent methods.
These gains are accompanied by stronger target-region attention, reduced hallucinations, and less forgetting on standard VQA, supporting our central claim: stronger visual grounding can elicit multimodal ICL without explicit ICL training.

\bibliographystyle{plainnat}
\bibliography{neurips_2026}

\newpage
\appendix

\section{Technical Appendices and Supplementary Material}

\subsection{Additional Analyses on Multimodal ICL Captioning Tasks}
\label{sec:appx:additional_ICL_captioning_results}
We first provide further analysis of Tab.~\ref{tab:icl_captioning_qwen}. 
Importantly, \oursbenchmark shows high zero-shot performance, even though correct answers require visual grounding of demonstrations in our proposed \oursbenchmark.
This is because we use ROUGE-L for captioning evaluation: even if the personalized identifier is incorrect, models can still achieve high scores by correctly describing generic visual content (\eg, actions or scenes).
Despite this limitation, ROUGE-L remains appropriate as captioning evaluation not only require evaluating correct identifier usage but also the overall quality of visual descriptions.

However, to isolate personalized identifier accuracy from overall description quality, we use GPT to evaluate whether generated captions correctly match the identifiers in the ground-truth captions. 
The evaluation prompt is as follows:

\begin{tcolorbox}[width=\linewidth, colframe=black, before skip=2mm, after skip=0.0cm, boxsep=0.0cm, middle=0.0cm, top=0.3cm, bottom=0.3cm]
\small
\ttfamily
You will be given: \\
1. A predicted caption \\
2. A ground-truth (GT) caption \\ \\ 

Your task is to identify the characters described in the GT caption and determine whether the predicted caption correctly assigns and uses the character names. \\ \\ 

Evaluation instructions: \\ \\

1. Carefully understand the GT caption: \\
- Identify all characters mentioned or implied. \\
- Identify which actions, attributes, clothing, positions, or interactions correspond to each character. \\ \\

2. Analyze the predicted caption: \\
- Determine which predicted character corresponds to which GT character. \\ 
- Check whether the predicted character names are correctly assigned. \\ \\

3. A prediction is considered incorrect if: \\ 
- A character is given the wrong name. \\
- Two characters’ names are swapped. \\ 
- A character from the GT caption is missing. \\ 
- An extra unsupported character name is introduced. \\

4. A prediction is considered correct only if: \\
- Every character identity matches the GT caption. \\
- The names are assigned to the correct corresponding characters. 
\end{tcolorbox}
\vspace{1em}

As shown in Tab.~\ref{tab:personalized_identifier}, \ours achieves the highest personalized identifier accuracy across all shots. Together with its ROUGE-L gains (Tab.~\ref{tab:icl_captioning_qwen}), this indicates that \ours better grounds personalized identifiers, leading to more accurate descriptions of both the target entity and other visual details.

Furthermore, we evaluate multimodal ICL captioning across diverse base MLLMs (Sec.~\ref{sec:appx:diverse_models}) and model sizes (Sec.~\ref{sec:appx:diverse_sizes}).

\paragraph{Human Validation of the GPT-4o-mini Identifier-Accuracy Metric.}
We perform a human-agreement study to validate that the GPT-4o-mini-based identifier-accuracy metric accurately evaluates personalized identifier binding. 
Specifically, we randomly sample 30 query examples and present each ground-truth/predicted caption pair to 50 independent human annotators through a questionnaire. 
Consistent with the GPT-4o-mini evaluation protocol, human evaluators are provided only with the ground-truth and predicted captions, without access to the corresponding query images. 
For each example, human evaluators answer the question: ``Are all personalized identifiers correctly assigned?'' with either Correct or Incorrect. 
We then evaluate the same caption pairs using our GPT-4o-mini-based identifier-accuracy metric and compare its predictions with the human judgments.

\begin{wraptable}{r}{0.48\textwidth}
\centering
\vspace{-1em}
\begin{tabular}{lcc}
\toprule
\textbf{Metric} 
& \textbf{Mean$\pm$SD} 
& \textbf{Median} \\
\midrule
Agreement (\%) 
& $95.91{\pm}2.98$ 
& $96.67$ \\

Cohen's $\kappa$ 
& $0.92{\pm}0.08$ 
& $0.93$ \\

Precision (\%) 
& $96.97{\pm}4.72$ 
& $100$ \\

Recall (\%) 
& $95.43{\pm}5.00$ 
& $100$ \\

F1-score (\%) 
& $95.97{\pm}3.88$ 
& $96.77$ \\

Raw score (/30) 
& $28.77{\pm}1.19$ 
& $29$ \\
\bottomrule
\end{tabular}
\caption{Human agreement study on personalized identifier accuracy.}
\label{tab:human_agreement}
\end{wraptable}

We summarize the results in Tab.~\ref{tab:human_agreement}, reporting (i) Agreement: the percentage of examples on which the GPT-4o-mini-based identifier-accuracy metric agrees with the majority human judgment, (ii) Cohen's $\kappa$: a standard measure of inter-rater agreement that quantifies the consistency between GPT-4o-mini and human judgments beyond chance agreement, and (iii) precision, recall, and F1-score. 
As shown, the GPT-4o-mini-based metric achieves high agreement with human judgments across all metrics, demonstrating that it provides a reliable evaluation of personalized identifier binding.

\begin{table}[t!]
\centering
\small
\resizebox{0.75\linewidth}{!}{
\begin{tabular}{llcccc}
    \toprule
    \multirow{2}{*}{\textbf{Category}} & \multirow{2}{*}{\textbf{Method}} & \multicolumn{4}{c}{\textbf{PICL-PororoSV}} \\ 
    & & 1-shot & 2-shot & 3-shot & 4-shot \\
    
    \cmidrule(lr){1-1} \cmidrule(lr){2-2} \cmidrule(lr){3-6}
    
    Off-the-shelf 
    & -
    & 38 & 21 & 29 & 31 \\
    
    \cmidrule(lr){1-1} \cmidrule(lr){2-2} \cmidrule(lr){3-6}
    
    \multirow{3}{*}{\makecell{Retrieval and \\ ordering}}
    & RICES {\footnotesize \color{blue}{(NeurIPS22)}}
    & 33 & 27 & 29 & 35 \\
    
    & VICL {\footnotesize \color{blue}{(ACL24)}}
    & 36 & 28 & 30 & 35 \\
    
    & MMICES {\footnotesize \color{blue}{(WACV25)}}
    & 30 & 24 & 27 & 32 \\
    
    \cmidrule(lr){1-1} \cmidrule(lr){2-2} \cmidrule(lr){3-6}
    
    \multirow{2}{*}{Task vector}
    & MTV {\footnotesize \color{blue}{(NeurIPS24)}}
    & 34 & 13 & 26 & 25 \\
    
    & FV {\footnotesize \color{blue}{(ICLR24)}}
    & 36 & 22 & 28 & 32 \\
    
    \cmidrule(lr){1-1} \cmidrule(lr){2-2} \cmidrule(lr){3-6}
    
    \multirow{4}{*}{ICL training}
    & MT-ICL {\footnotesize \color{blue}{(ECCVW24)}}
    & 54 & \underline{54} & \underline{56} & \textbf{59} \\
    
    & DARA {\footnotesize \color{blue}{(COLM25)}}
    & 43 & 37 & 41 & 47 \\
    
    & MimIC {\footnotesize \color{blue}{(CVPR25)}}
    & 51 & 43 & 44 & 49 \\
    
    & STV {\footnotesize \color{blue}{(AAAI26)}}
    & \underline{59} & 52 & 55 & 56 \\
    
    \cmidrule(lr){1-1} \cmidrule(lr){2-2} \cmidrule(lr){3-6}
    
    \multirow{7}{*}{\makecell{Mitigating \\ hallucination}}
    & PAI {\footnotesize \color{blue}{(ECCV24)}}
    & 26 & 17 & 25 & 27 \\
    
    & VOLCANO {\footnotesize \color{blue}{(NAACL24)}}
    & 54 & 41 & 47 & 55 \\
    
    & LURE {\footnotesize \color{blue}{(ICLR24)}}
    & 24 & 19 & 24 & 28 \\
    
    & LRV {\footnotesize \color{blue}{(ICLR24)}}
    & 42 & 29 & 25 & 34 \\
    
    & IBD {\footnotesize \color{blue}{(CVPRW25)}}
    & 50 & 45 & 46 & 42 \\
    
    & SPARC {\footnotesize \color{blue}{(ICML25)}}
    & 54 & 42 & 51 & 55 \\
    
    & \ours (Ours)
    & \textbf{62} & \textbf{57} & \textbf{57} & \textbf{59} \\
    
    \bottomrule
    \end{tabular}
    }
\vspace{0.5em}
\caption{
    \textbf{Comparison of personalized identifier accuracy.}
    We measure whether predicted captions correctly match personalized identifiers in the ground-truth captions for each query.
    \ours achieves the highest accuracy across all shots.
}
\label{tab:personalized_identifier}
\end{table}

\subsection{Details of Experimental Setups}
\label{sec:appx:exp_details}


\paragraph{Metrics.}
To quantify visual grounding, we examine whether MLLMs attend to relevant regions during token generation. 
To this end, we use target attention score (TAS)~\cite{ou2026mllms} measures the average attention weight assigned across all attention heads from each generated token to image tokens corresponding to a target region, where the region is defined by a segmentation map obtained from SAM3~\cite{carion2025sam}.
Formally, let $\alpha^{(h)}_{t_i, t_g}$ denote the attention weight from the generated token $t_g$ to image token $t_i$ at attention head $h$, and let $\mathcal{T}_\text{target}$ denote the set of image tokens corresponding to the target region associated with $t_g$.
Then, TAS $a$ is defined as follows:
\begin{equation}
    a = \frac{1}{H} \sum_{h=1}^{H}\frac{1}{|\mathcal{T}_\text{target}|}\sum_{t_i \in \mathcal{T}_\text{target}}\alpha^{(h)}_{t_i, t_g}.
\end{equation}

Moreover, to quantitatively evaluate visual hallucination, we use the OpenCHAIR metric~\cite{ben2024mitigating}, which decomposes predicted captions into words and measures the proportion of hallucinated words by checking whether each predicted word is supported by the ground-truth caption, considering synonyms through LLMs.

\paragraph{Implementation Details and Hyperparameters.}
We set the batch size to 16, the learning rate for Qwen-VL-3 and LLaVA-OneVision-1.5 to $2 \times 10^{-5}$, and InternVL to $2 \times 10^{-6}$, the Cosine LR scheduler and the Adam optimizer~\citep{loshchilov2019decoupled}, and LoRA rank $r$ of 128 attached to all linear layers in the LLM backbone for all baselines require training, such as ICL data-dependent baselines and hallucination mitigation baselines (including \ours).
For \ours, we set the reward weighting coefficients as $\lambda_I=2$, $\lambda_T=1$, and $\lambda_s=4$ in all experiments. 
See Sec.~\ref{sec:appx:hyper_analysis} for detailed hyperparameter analysis.
All experiments are executed in Python 3.10, on four Ubuntu 20.04 machines, with 8 NVIDIA RTX A6000 GPUs each. 
Each experiment runs on a single RTX A6000 GPU.
We report single-run results, since repeated ICL with the same inputs yields low variance in MLLM outputs~\cite{alayrac2022flamingo, luo2024does}.
For trainable baselines, we use 1,000 training samples for fair comparison, following the training datasets used in each method’s original paper: ICL-dependent methods sample from the evaluated ICL benchmark; LURE uses MS COCO~\cite{lin2014microsoft}, VOLCANO uses LLaVA-SFT-127k~\cite{liu2023visual}, LRV uses Visual Genome~\cite{krishna2017visual}, and \ours uses 1,000 samples from each of InpaintCOCO~\cite{rosch2024enhancing} and EQBEN~\cite{wang2023equivariant}.

\subsection{Hyperparameter Analysis}
\label{sec:appx:hyper_analysis}

We analyze the effect of reward weights $\lambda_I$, $\lambda_T$, and $\lambda_s$, which weight the image contrastive reward $R_I^{(p)}$, text contrastive reward $R_T^{(p)}$, and shared-content reward $R_s^{(p)}$, respectively.
Since these coefficients control relative contributions of each reward term, their ratios, rather than absolute values, are key. 
Therefore, we fix one weight $\lambda_s$ and vary the others (\ie, $\lambda_I$, $\lambda_T$) to study the effect of different relative ratios.

\begin{figure}[t!]
\centering

\begin{subfigure}{0.32\linewidth}
    \centering
    \includegraphics[width=\linewidth]{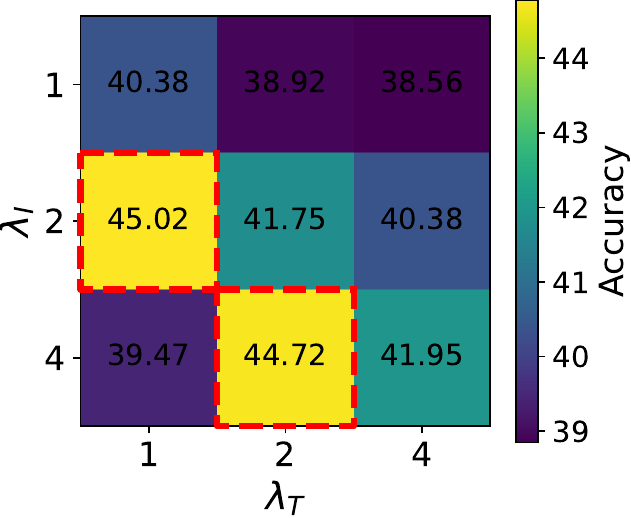}
    \caption{1-shot ICL Accuracy}
\end{subfigure}
\hfill
\begin{subfigure}{0.32\linewidth}
    \centering
    \includegraphics[width=\linewidth]{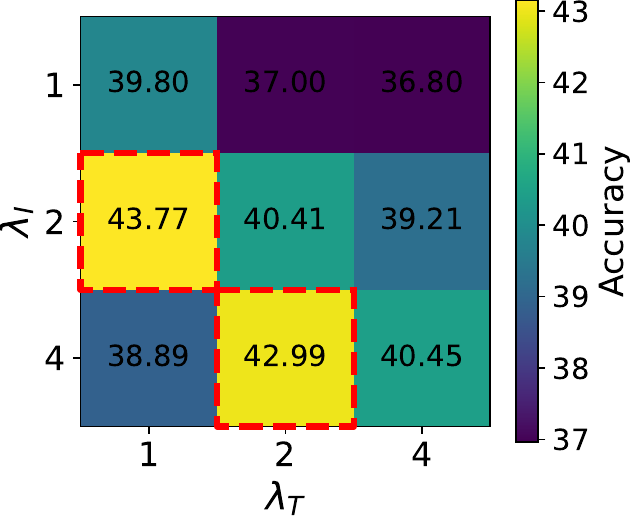}
    \caption{4-shot ICL Accuracy}
\end{subfigure}
\hfill
\begin{subfigure}{0.32\linewidth}
    \centering
    \includegraphics[width=\linewidth]{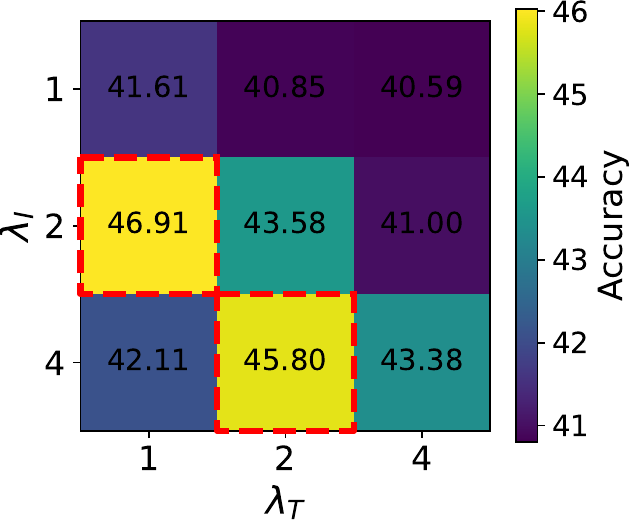}
    \caption{8-shot ICL Accuracy}
\end{subfigure}

\caption{
    \textbf{Accuracy heatmap over reward weights ($\lambda_I$, $\lambda_T$) with $\lambda_s=4$.}
    If the text reward weight ($\lambda_T$) outweighs the visual contrastive reward weight ($\lambda_I$), the model over-focuses on textual demonstrations, weakening visual grounding and degrading performance. 
    Conversely, if $\lambda_I$ is too large compared to $\lambda_T$, the model underutilizes useful information from ground-truth captions, also leading to performance drops.
    Moderate ratios that emphasize visual rewards while retaining text signals (\eg, $\lambda_I/\lambda_T=2$, red dotted box) consistently show strong performance across shots.
    We evaluate ICL performance on the MS COCO captioning benchmark, with InpaintCOCO used for contrastive RL training in \ours.
}
\label{fig:hyperparameter_analysis}
\end{figure}

As shown in Fig.~\ref{fig:hyperparameter_analysis}, placing more weight on the text contrastive reward than the image contrastive reward (\ie, $\lambda_T > \lambda_I$, strictly upper triangle in the heatmap) leads to lower performance.
This is because it may bias the reward toward over-reliance on ground-truth captions in contrastive pairs, encouraging simple imitation of the corresponding caption rather than enhancing visual grounding.
Conversely, overly emphasizing the image contrastive reward (\ie, $\lambda_I=4, \lambda_T=1$) can underutilize useful details from ground-truth descriptions, also leading to performance drops.
Considering this trade-off and our goal of improving visual grounding, we set $\lambda_I > \lambda_T$ while keeping their ratio moderate to avoid overemphasis.
Specifically, we use $\lambda_I=2$, $\lambda_T=1$, and $\lambda_s=4$ across all experiments.

\subsection{Additional Details and Analysis of Proposition Decomposition}
\label{sec:appx:detail_prompt_decomposition}

\paragraph{Detailed Procedure.}
Proposition decomposition with GPT-4o-mini is performed online during RL training. More concretely, at each GRPO step, the MLLM first generates captions for the contrastive image pair. We then send the generated captions together with the corresponding ground-truth captions to the GPT-4o-mini API to decompose them into propositions.

\paragraph{Computational Cost.}
Importantly, the decomposition operates only on textual captions (\ie, the generated and ground-truth captions), without requiring any image inputs, resulting in minimal computational and API overhead. The additional API cost is approximately $4 \times 10^{-4}$ USD per training sample, while the wall-clock training time increases by only 5.32\%. 
This modest overhead is because the API calls are lightweight text-only requests and can be executed in parallel with GRPO rollout generation using multiple threads. 

\paragraph{Alternative Proposition Decomposers.}
We emphasize that proposition decomposition is a lightweight text-only preprocessing step that simply partitions the generated and ground-truth captions into unique and shared propositions. 
Therefore, GPT-4o-mini is not essential to our method and can be readily replaced by much smaller models with comparable performance. 
As shown in Tab.~\ref{tab:decomposer_picl} and Tab.~\ref{tab:decomposer_truemicl}, small open-source LLMs (\eg, Qwen3-1.7B, LLaMA3.2-3B) achieve comparable performance. 
This demonstrates that the performance gains stem from the proposition decomposition itself, rather than from the high capability of GPT-4o-mini.

\begin{table}[ht]
\centering
\begin{tabular}{lcccc}
\toprule
\textbf{Proposition Decomposer}
& \textbf{0-shot}
& \textbf{1-shot}
& \textbf{2-shot}
& \textbf{4-shot} \\
\midrule
GPT
& $33.04{\pm}1.52$
& $35.82{\pm}1.44$
& $35.31{\pm}1.88$
& $35.30{\pm}1.19$ \\

Qwen3-1.7B
& $32.55{\pm}1.30$
& $32.74{\pm}2.30$
& $36.74{\pm}1.29$
& $35.20{\pm}1.66$ \\

LLaMA3.2-3B
& $32.32{\pm}1.41$
& $34.44{\pm}1.72$
& $36.87{\pm}1.66$
& $37.60{\pm}2.15$ \\

Qwen3-4B
& $31.88{\pm}1.14$
& $34.97{\pm}1.87$
& $37.62{\pm}2.05$
& $34.41{\pm}0.96$ \\
\bottomrule
\end{tabular}
\vspace{0.5em}
\caption{Comparison of different LLM-based proposition decomposers in ProCoRe on PICL.}
\label{tab:decomposer_picl}
\end{table}

\begin{table}[ht]
\centering
\begin{tabular}{lcccc}
\toprule
\textbf{Proposition Decomposer}
& \textbf{0-shot}
& \textbf{1-shot}
& \textbf{2-shot}
& \textbf{4-shot} \\
\midrule
GPT
& $32.44{\pm}1.89$
& $50.91{\pm}2.74$
& $55.44{\pm}1.59$
& $61.37{\pm}1.68$ \\

Qwen3-1.7B
& $31.98{\pm}1.79$
& $53.90{\pm}1.34$
& $54.04{\pm}1.34$
& $63.73{\pm}1.45$ \\

LLaMA3.2-3B
& $31.32{\pm}1.72$
& $50.02{\pm}1.56$
& $53.33{\pm}1.60$
& $60.90{\pm}1.69$ \\

Qwen3-4B
& $33.47{\pm}1.42$
& $50.26{\pm}2.18$
& $54.91{\pm}0.89$
& $62.38{\pm}1.67$ \\
\bottomrule
\end{tabular}
\vspace{0.5em}
\caption{Comparison of different LLM-based proposition decomposers in ProCoRe on TrueMICL.}
\label{tab:decomposer_truemicl}
\end{table}

\subsection{Detailed ablation study of \ours's contrastive reward}
\label{sec:appx:detailed_ablation}
We conduct detailed ablation studies with the following variants: (i) Image contrastive only (\ie, $\lambda_T=0, \lambda_s=0$ in Eq.~\ref{eq:objective}), (ii) Text contrastive only (\ie, $\lambda_I=0, \lambda_s=0$ in Eq.~\ref{eq:objective}), (iii) No shared-content reward (\ie, $\lambda_s=0$) and (iv) No proposition decomposition (\ie, sentence-level contrastive reward). 
We summarize the results in Tab.~\ref{tab:detailed_ablation_picl} and Tab.~\ref{tab:detailed_ablation_truemicl}.
As shown, using either the contrastive image reward or the contrastive text reward alone yields only marginal improvements over the off-the-shelf model, while combining both rewards leads to substantially better performance. 
These results indicate that jointly encouraging the model to attend to discriminative visual evidence and to generate semantically faithful descriptions is important for improving multimodal ICL.

Furthermore, incorporating the shared-content reward consistently improves performance, suggesting that explicitly rewarding propositions shared across the contrastive pair prevents the model from focusing solely on discriminative differences and encourages it to capture visual attributes common to both images. 
Finally, replacing proposition-level decomposition with a sentence-level contrastive reward leads to a noticeable performance drop, highlighting the importance of fine-grained proposition-level supervision. 
Together, these ablations demonstrate the complementary contributions of the individual reward components and provide further support for the design of our proposition-level contrastive reward.

\begin{table*}[ht]
\centering
\begin{tabular}{lcccc}
\toprule
\textbf{Method} 
& \textbf{1-shot} 
& \textbf{2-shot} 
& \textbf{3-shot} 
& \textbf{4-shot} \\
\midrule
Off-the-shelf 
& 23.33 & 27.00 & 31.67 & 29.00 \\

SFT 
& 9.67 & 9.33 & 15.67 & 10.33 \\

No-proposition-decomposition
& 42.67 & 38.67 & 40.00 & 32.00 \\

\midrule

Image contrastive only ($\lambda_T = 0, \lambda_s = 0$)
& 40.00 & 42.67 & 37.67 & 30.33 \\

Text contrastive only ($\lambda_I = 0, \lambda_s = 0$) 
& 33.00 & 30.33 & 31.67 & 32.00 \\

No shared-content reward ($\lambda_s = 0$)
& 53.33 & 38.00 & 44.00 & 39.67 \\

\midrule
\textbf{ProCoRe (Ours)} 
& \textbf{62.67} 
& \textbf{58.67} 
& \textbf{53.00} 
& \textbf{50.33} \\
\bottomrule
\end{tabular}
\caption{\textbf{Ablation study of ProCoRe's individual reward components on PICL using personalized identifier accuracy.}}
\label{tab:detailed_ablation_picl}
\end{table*}

\begin{table*}[ht]
\centering
\begin{tabular}{lccccc}
\toprule
\textbf{Method} 
& \textbf{0-shot} 
& \textbf{1-shot} 
& \textbf{2-shot} 
& \textbf{4-shot} 
& \textbf{8-shot} \\
\midrule
Off-the-shelf 
& 15.53 & 19.02 & 21.21 & 23.52 & 25.65 \\

SFT 
& 19.48 & 27.92 & 30.11 & 26.84 & 37.62 \\

No-proposition-decomposition
& 36.02 & 44.73 & 43.21 & 44.11 & 42.58 \\

\midrule

Image contrastive only ($\lambda_T = 0, \lambda_s = 0$)
& 22.91 & 35.44 & 29.37 & 33.88 & 31.26 \\

Text contrastive only ($\lambda_I = 0, \lambda_s = 0$) 
& 24.12 & 36.91 & 27.54 & 34.26 & 28.93 \\

No shared-content reward ($\lambda_s = 0$)
& 30.06 & 30.58 & 30.11 & 35.49 & 37.82 \\

\midrule
\textbf{ProCoRe (Ours)} 
& \textbf{39.12} 
& \textbf{46.88} 
& \textbf{46.41} 
& \textbf{46.26} 
& \textbf{46.03} \\
\bottomrule
\end{tabular}
\caption{\textbf{Ablation study of ProCoRe's individual reward components on TrueMICL using ROUGE-L.}}
\label{tab:detailed_ablation_truemicl}
\end{table*}

\subsection{Additional Experiments on Multimodal ICL Classification Tasks}
\label{sec:appx:additional_ICL_classification_results}

In addition to TrueMICL and ISEKAI, we also evaluate on MIC~\cite{zhao2024mmicl} and VL-ICL~\cite{zong2025vlicl}, and summarize the results in Tab.~\ref{tab:icl_qwen_vlicl_mmicl}.
ICL data-dependent baselines (\eg, task-vector and ICL-training methods) sometimes outperform \ours on the benchmarks used for their ICL training or extraction, but degrade drastically on others, indicating poor generalization.

In contrast, despite using no ICL data, \ours matches or even exceeds these baselines on their training benchmarks and significantly outperforms them on unseen ones. 
This supports our hypothesis that improving visual grounding yields broadly generalizable multimodal ICL gains, rather than benchmark-specific improvements.

\begin{table}[t!]
\vspace{-0.5em}
\centering
\small
\resizebox{\linewidth}{!}{
\begin{tabular}{lclcccccccccc}
\toprule
\multirow{2}{*}{\textbf{Category}} 
& \multirow{2}{*}{\makecell{\textbf{ICL Training}\\ \textbf{Dataset}}} 
& \multirow{2}{*}{\textbf{Method}} 
& \multicolumn{5}{c}{\textbf{VL-ICL}} 
& \multicolumn{5}{c}{\textbf{MIC}} \\

& & 
& 0-shot & 1-shot & 2-shot & 4-shot & 8-shot
& 0-shot & 1-shot & 2-shot & 4-shot & 8-shot \\

\cmidrule(lr){1-1} 
\cmidrule(lr){2-2} 
\cmidrule(lr){3-3} 
\cmidrule(lr){4-8} 
\cmidrule(lr){9-13}

Off-the-shelf 
& \multirow{1}{*}{-}
& -
& 19.29 & 45.13 & 59.25 & 57.42 & 63.25
& 16.49 & 24.90 & 30.23 & 35.48 & 42.46 \\

\cmidrule(lr){1-1} 
\cmidrule(lr){2-2} 
\cmidrule(lr){3-3} 
\cmidrule(lr){4-8} 
\cmidrule(lr){9-13}

\multirow{3}{*}{\makecell{Retrieval and \\ ordering}}
& \multirow{3}{*}{-}
& RICES {\footnotesize \color{blue}{(NeurIPS22)}}
& 19.29 & 45.13 & 44.67 & 53.13 & 57.88
& 16.49 & 24.90 & 30.70 & 37.32 & 45.48 \\

& & VICL {\footnotesize \color{blue}{(ACL24)}}
& 15.84 & 41.72 & 41.03 & 53.68 & 58.21
& 14.11 & 27.42 & 32.88 & 33.21 & 45.03 \\

& & MMICES {\footnotesize \color{blue}{(WACV25)}}
& 19.29 & 45.13 & 44.08 & 53.33 & 59.84
& 16.49 & 24.90 & 30.68 & 36.62 & 41.94 \\

\cmidrule(lr){1-1} 
\cmidrule(lr){2-2} 
\cmidrule(lr){3-3} 
\cmidrule(lr){4-8} 
\cmidrule(lr){9-13}

\multirow{4}{*}{Task vector}
& \multirow{2}{*}{VL-ICL}
& MTV {\footnotesize \color{blue}{(NeurIPS24)}}
& 20.96 & 37.79 & 38.92 & 44.04 & 48.88
& 33.74 & 35.21 & 36.02 & 37.88 & 38.41 \\

& & FV {\footnotesize \color{blue}{(ICLR24)}}
& 20.11 & 46.02 & 60.47 & 56.88 & 64.73
& 14.11 & 22.03 & 26.18 & 30.12 & 34.27 \\

\cmidrule(lr){2-2} 
\cmidrule(lr){3-3} 
\cmidrule(lr){4-8} 
\cmidrule(lr){9-13}

& \multirow{2}{*}{MIC}
& MTV {\footnotesize \color{blue}{(NeurIPS24)}}
& 16.42 & 36.11 & 34.28 & 42.63 & 47.21 
& 38.14 & 39.97 & 40.46 & 41.99 & 42.95 \\

& & FV {\footnotesize \color{blue}{(ICLR24)}}
& 15.92 & 41.18 & 55.84 & 52.11 & 59.27
& 18.78 & 26.77 & 30.97 & 34.69 & 39.38 \\

\cmidrule(lr){1-1} 
\cmidrule(lr){2-2} 
\cmidrule(lr){3-3} 
\cmidrule(lr){4-8} 
\cmidrule(lr){9-13}

\multirow{8}{*}{ICL training}
& \multirow{4}{*}{VL-ICL}
& MT-ICL {\footnotesize \color{blue}{(ECCVW24)}}
& \underline{23.18} & 47.92 & 48.63 & 49.11 & 50.02
& 35.42 & 32.03 & 31.88 & 34.01 & 43.92 \\

& & DARA {\footnotesize \color{blue}{(COLM25)}}
& 19.29 & 47.27 & \underline{61.41} & 56.06 & 64.72
& 16.49 & 21.88 & 24.17 & 29.04 & 38.11 \\

& & MimIC {\footnotesize \color{blue}{(CVPR25)}}
& 23.61 & 46.55 & 60.12 & \textbf{62.48} & 64.09
& 22.88 & 19.21 & 20.11 & 21.04 & 35.18 \\

& & STV {\footnotesize \color{blue}{(AAAI26)}}
& 21.07 & \underline{48.63} & 59.77 & 59.34 & \textbf{68.11}
& 36.41 & 33.18 & 32.74 & 34.21 & 43.03 \\

\cmidrule(lr){2-2} 
\cmidrule(lr){3-3} 
\cmidrule(lr){4-8} 
\cmidrule(lr){9-13}

& \multirow{4}{*}{MIC}
& MT-ICL {\footnotesize \color{blue}{(ECCVW24)}}
& 20.41 & 32.88 & 49.21 & 47.63 & 54.02
& 49.01 & 54.41 & 55.18 & 55.28 & \underline{55.84} \\

& & DARA {\footnotesize \color{blue}{(COLM25)}}
& 17.84 & 34.11 & 45.73 & 41.92 & 50.88 
& 16.49 & 44.18 & 46.72 & 41.63 & 49.91 \\

& & MimIC {\footnotesize \color{blue}{(CVPR25)}}
& 22.03 & 36.74 & 50.11 & 46.28 & 53.42
& 36.02 & 41.92 & 42.03 & 42.85 & 46.38 \\

& & STV {\footnotesize \color{blue}{(AAAI26)}}
& 19.18 & 37.21 & 47.88 & 48.03 & 56.62
& \underline{49.52} & \underline{55.14} & \textbf{55.30} & \underline{55.53} & 55.38 \\

\cmidrule(lr){1-1} 
\cmidrule(lr){2-2} 
\cmidrule(lr){3-3} 
\cmidrule(lr){4-8} 
\cmidrule(lr){9-13}

\multirow{7}{*}{\makecell{Mitigating \\ hallucination}}
& \multirow{7}{*}{-}
& PAI {\footnotesize \color{blue}{(ECCV24)}}
& 15.94 & 40.88 & 55.73 & 53.62 & 59.11
& 30.18 & 35.41 & 36.74 & 37.92 & 40.11 \\

& & VOLCANO {\footnotesize \color{blue}{(NAACL24)}} 
& 20.88 & 46.72 & 60.41 & 58.03 & 65.74
& 38.74 & 44.11 & 39.88 & 46.21 & 49.03 \\

& & LURE {\footnotesize \color{blue}{(ICLR24)}} 
& 17.02 & 42.51 & 54.88 & 52.77 & 60.42
& 34.18 & 39.74 & 43.92 & 41.27 & 45.11 \\

& & LRV {\footnotesize \color{blue}{(ICLR24)}}
& 18.41 & 47.09 & 61.02 & 59.11 & 64.38
& 33.41 & 39.12 & 45.28 & 40.74 & 43.91 \\

& & IBD {\footnotesize \color{blue}{(CVPRW25)}} 
& 18.37 & 46.44 & 60.03 & 55.88 & 64.71
& 39.27 & 43.83 & 40.11 & 45.63 & 47.88 \\

& & SPARC {\footnotesize \color{blue}{(ICML25)}} 
& 21.67 & 44.58 & 58.12 & 56.04 & 66.03
& 32.18 & 38.74 & 44.62 & 41.92 & 50.21 \\

& & \ours
& \textbf{25.91} & \textbf{49.88} & \textbf{62.54} & \underline{61.47} & \underline{67.03}
& \textbf{52.11} & \textbf{58.29} & \underline{53.88} & \textbf{58.01} & \textbf{56.63} \\

\bottomrule
\end{tabular}
}
\vspace{0.4em}
\caption{
\textbf{Quantitative comparison on the VL-ICL benchmark and MIC.}
Best results for each shot are highlighted in bold, while second-best results are underlined.
All methods use Qwen-VL-3-8B as the base MLLM.
Demonstrations are randomly ordered for all methods except retrieval- and ordering-based approaches.
}
\label{tab:icl_qwen_vlicl_mmicl}
\end{table}

\subsection{Quantitative Analysis across Diverse MLLMs}
\label{sec:appx:diverse_models}
Beyond Qwen-VL-3, we also evaluate \ours with LLaVA-OneVision-1.5-8B~\cite{an2025llava} and InternVL-3.5-8B~\cite{wang2025internvl3} to demonstrate robustness across MLLMs.
As shown in Tab.~\ref{tab:icl_captioning_onevision_combined} and Tab.~\ref{tab:icl_captioning_internvl_combined}, \ours consistently outperforms baselines, demonstrating strong cross-model applicability.

\begin{table}[t!]
\centering
\small
\resizebox{\linewidth}{!}{
\begin{tabular}{lclcccccccccc}
\toprule
\multirow{2}{*}{\textbf{Category}} 
& \multirow{2}{*}{\makecell{\textbf{ICL Training}\\ \textbf{Dataset}}} 
& \multirow{2}{*}{\textbf{Method}} 
& \multicolumn{5}{c}{\textbf{MS COCO}} 
& \multicolumn{5}{c}{\textbf{\oursbenchmark}} \\

& & 
& 0-shot & 1-shot & 2-shot & 4-shot & 8-shot 
& 0-shot & 1-shot & 2-shot & 3-shot & 4-shot \\

\cmidrule(lr){1-1} \cmidrule(lr){2-2} \cmidrule(lr){3-3}
\cmidrule(lr){4-8} \cmidrule(lr){9-13}

Off-the-shelf 
& - 
& Random
& 26.92 & 31.22 & 33.25 & 35.83 & 36.26
& 18.44 & 21.86 & 24.56 & 25.84 & 28.01 \\

\cmidrule(lr){1-1}\cmidrule(lr){2-2}\cmidrule(lr){3-3}
\cmidrule(lr){4-8}\cmidrule(lr){9-13}

\multirow{3}{*}{\makecell{Retrieval and \\ ordering}}
& \multirow{3}{*}{-}

& RICES {\footnotesize \color{blue}{(NeurIPS22)}}
& 26.92 & 31.22 & 31.84 & 33.70 & 35.41
& 18.44 & 21.86 & 25.01 & 26.31 & 27.08 \\

&
& VICL {\footnotesize \color{blue}{(ACL24)}}
& 27.14 & 32.81 & 35.02 & 37.06 & 36.41
& 18.72 & 23.11 & 23.56 & 24.84 & 27.01 \\

&
& MMICES {\footnotesize \color{blue}{(WACV25)}}
& 26.92 & 31.22 & 29.62 & 33.71 & 36.65
& 18.44 & 21.86 & 20.01 & 20.99 & 23.31 \\

\cmidrule(lr){1-1}\cmidrule(lr){2-2}\cmidrule(lr){3-3}
\cmidrule(lr){4-8}\cmidrule(lr){9-13}

\multirow{4}{*}{Task vector}

& \multirow{2}{*}{MS COCO}
& MTV {\footnotesize \color{blue}{(NeurIPS24)}}
& 22.56 & 22.75 & 24.40 & 25.00 & 26.63
& 9.84 & 13.72 & 12.06 & 12.88 & 11.91 \\

&
& FV {\footnotesize \color{blue}{(ICLR24)}}
& 26.00 & 31.29 & 32.18 & 32.94 & 36.58
& 17.42 & 21.08 & 21.76 & 20.94 & 24.71 \\

\cmidrule(lr){2-2}\cmidrule(lr){3-3}
\cmidrule(lr){4-8}\cmidrule(lr){9-13}

& \multirow{2}{*}{\oursbenchmark}
& MTV {\footnotesize \color{blue}{(NeurIPS24)}}
& 20.87 & 21.12 & 23.18 & 23.74 & 25.41
& 11.56 & 15.19 & 14.25 & 14.61 & 13.53 \\

&
& FV {\footnotesize \color{blue}{(ICLR24)}}
& 24.31 & 29.48 & 30.21 & 31.02 & 34.26
& 19.53 & 22.81 & 23.49 & 23.03 & 26.33 \\

\cmidrule(lr){1-1}\cmidrule(lr){2-2}\cmidrule(lr){3-3}
\cmidrule(lr){4-8}\cmidrule(lr){9-13}

\multirow{8}{*}{ICL training}

& \multirow{4}{*}{MS COCO}
& MT-ICL {\footnotesize \color{blue}{(ECCVW24)}}
& 41.39 & \underline{45.18} & \underline{44.56} & \underline{45.64} & \underline{43.19}
& 16.12 & 19.73 & 21.04 & 21.58 & 23.41 \\

&
& DARA {\footnotesize \color{blue}{(COLM25)}}
& 26.92 & 32.50 & 34.52 & 35.29 & 37.66
& 16.35 & 19.42 & 21.18 & 22.74 & 24.63 \\

&
& MimIC {\footnotesize \color{blue}{(CVPR25)}}
& 26.73 & 29.64 & 33.95 & 34.86 & 37.88
& 17.23 & 20.16 & 22.38 & 22.06 & 24.72 \\

&
& STV {\footnotesize \color{blue}{(AAAI26)}}
& \underline{42.52} & 42.21 & 43.34 & 43.51 & 42.53
& 14.81 & 21.56 & 22.17 & 22.84 & 23.09 \\

\cmidrule(lr){2-2}\cmidrule(lr){3-3}
\cmidrule(lr){4-8}\cmidrule(lr){9-13}

& \multirow{4}{*}{\oursbenchmark}
& MT-ICL {\footnotesize \color{blue}{(ECCVW24)}}
& 28.04 & 32.17 & 31.28 & 32.36 & 39.02
& 18.44 & 21.86 & 23.38 & 23.69 & 25.66 \\

&
& DARA {\footnotesize \color{blue}{(COLM25)}}
& 24.48 & 29.12 & 30.67 & 32.06 & 34.19
& 18.44 & 21.37 & 23.37 & 24.67 & 26.06 \\

&
& MimIC {\footnotesize \color{blue}{(CVPR25)}}
& 23.11 & 26.32 & 30.24 & 31.15 & 34.02
& 19.31 & 21.97 & 24.66 & 24.21 & 26.27 \\

&
& STV {\footnotesize \color{blue}{(AAAI26)}}
& 28.64 & 28.98 & 29.87 & 30.05 & 38.81
& \underline{24.27} & 23.43 & 24.24 & 24.36 & 25.01 \\

\cmidrule(lr){1-1}\cmidrule(lr){2-2}\cmidrule(lr){3-3}
\cmidrule(lr){4-8}\cmidrule(lr){9-13}

\multirow{7}{*}{\makecell{Mitigating \\ hallucination}}
& \multirow{7}{*}{-}

& PAI {\footnotesize \color{blue}{(ECCV24)}}
& 22.87 & 28.92 & 29.08 & 31.74 & 34.21
& 15.83 & 18.92 & 22.01 & 23.14 & 24.76 \\

&
& VOLCANO {\footnotesize \color{blue}{(NAACL24)}}
& 28.21 & 34.44 & 36.02 & 37.88 & 39.37
& \underline{21.02} & \underline{24.47} & \underline{27.89} & \underline{28.21} & \underline{30.04} \\

&
& LURE {\footnotesize \color{blue}{(ICLR24)}}
& 30.41 & 35.98 & 38.61 & 39.44 & 41.72
& 16.72 & 20.11 & 23.58 & 24.79 & 26.04 \\

&
& LRV {\footnotesize \color{blue}{(ICLR24)}}
& 27.05 & 32.62 & 35.11 & 36.47 & 38.02
& 17.21 & 20.34 & 24.12 & 25.08 & 26.02 \\

&
& IBD {\footnotesize \color{blue}{(CVPRW25)}}
& 26.88 & 33.16 & 34.73 & 37.02 & 39.05
& 17.87 & 23.11 & 26.42 & 27.05 & 28.66 \\

&
& SPARC {\footnotesize \color{blue}{(ICML25)}}
& 29.67 & 35.21 & 37.54 & 38.73 & 40.81
& 16.95 & 19.73 & 23.56 & 24.67 & 25.41 \\

&
& \ours (Ours)
& \textbf{43.22} & \textbf{47.91} & \textbf{46.04} & \textbf{46.21} & \textbf{45.23}
& \textbf{25.36} & \textbf{27.81} & \textbf{29.22} & \textbf{31.54} & \textbf{33.67} \\

\bottomrule
\end{tabular}
}
\vspace{0.5em}
\caption{
\textbf{Quantitative comparison in multimodal ICL captioning benchmarks.}
For each dataset and shot, the top scores are highlighted in bold, and the second-best scores are underlined.
We use LLaVA-OneVision-1.5-8B as the MLLM and apply random ordering of demonstrations for all baselines except retrieval- and ordering-based methods.
}
\vspace{-1em}
\label{tab:icl_captioning_onevision_combined}
\end{table}

\begin{table}[t!]
\centering
\small
\resizebox{\linewidth}{!}{
\begin{tabular}{lclcccccccccc}
\toprule
\multirow{2}{*}{\textbf{Category}} 
& \multirow{2}{*}{\makecell{\textbf{ICL Training}\\ \textbf{Dataset}}} 
& \multirow{2}{*}{\textbf{Method}} 
& \multicolumn{5}{c}{\textbf{MS COCO}} 
& \multicolumn{5}{c}{\textbf{\oursbenchmark}} \\ 
& & 
& 0-shot & 1-shot & 2-shot & 4-shot & 8-shot 
& 0-shot & 1-shot & 2-shot & 3-shot & 4-shot \\

\cmidrule(lr){1-1} \cmidrule(lr){2-2} \cmidrule(lr){3-3}
\cmidrule(lr){4-8} \cmidrule(lr){9-13}

Off-the-shelf & - & Random
& 27.57 & 31.78 & 33.47 & 34.11 & 33.99
& 15.68 & 14.58 & 17.37 & 16.82 & 19.95 \\

\cmidrule(lr){1-1} \cmidrule(lr){2-2} \cmidrule(lr){3-3}
\cmidrule(lr){4-8} \cmidrule(lr){9-13}

\multirow{3}{*}{\makecell{Retrieval and \\ ordering}} 
& \multirow{3}{*}{-}
& RICES {\footnotesize \color{blue}{(NeurIPS22)}}
& 27.57 & 31.78 & 33.29 & 33.05 & 34.06
& 15.68 & 14.58 & 18.27 & 22.19 & 25.02 \\

& & VICL {\footnotesize \color{blue}{(ACL24)}}
& 26.01 & 33.02 & 34.56 & 34.73 & 35.21
& 16.41 & 15.92 & 15.36 & 21.88 & 21.47 \\

& & MMICES {\footnotesize \color{blue}{(WACV25)}}
& 27.57 & 31.78 & 33.48 & 33.83 & 33.74
& 15.68 & 14.58 & 14.85 & 20.54 & 20.56 \\

\cmidrule(lr){1-1} \cmidrule(lr){2-2} \cmidrule(lr){3-3}
\cmidrule(lr){4-8} \cmidrule(lr){9-13}

\multirow{4}{*}{Task vector} 
& \multirow{2}{*}{MS COCO}
& MTV {\footnotesize \color{blue}{(NeurIPS24)}}
& 28.41 & 32.67 & 34.21 & 34.11 & 35.72
& 13.02 & 13.74 & 15.26 & 13.18 & 16.72 \\

& & FV {\footnotesize \color{blue}{(ICLR24)}}
& 27.98 & 33.02 & 34.05 & 33.64 & 35.18
& 14.11 & 12.92 & 15.21 & 14.03 & 17.48 \\

\cmidrule(lr){2-2} \cmidrule(lr){3-3}
\cmidrule(lr){4-8} \cmidrule(lr){9-13}

& \multirow{2}{*}{\oursbenchmark}
& MTV {\footnotesize \color{blue}{(NeurIPS24)}}
& 26.03 & 30.11 & 31.84 & 32.76 & 34.02
& 15.27 & 15.48 & 17.02 & 14.63 & 18.41 \\

& & FV {\footnotesize \color{blue}{(ICLR24)}}
& 25.74 & 31.18 & 32.41 & 31.92 & 33.47
& 16.08 & 14.36 & 17.68 & 15.85 & 19.05 \\

\cmidrule(lr){1-1} \cmidrule(lr){2-2} \cmidrule(lr){3-3}
\cmidrule(lr){4-8} \cmidrule(lr){9-13}

\multirow{8}{*}{ICL training} 
& \multirow{4}{*}{MS COCO}
& MT-ICL {\footnotesize \color{blue}{(ECCVW24)}}
& \underline{36.63} & \underline{37.91} & \underline{38.11} & \underline{38.30} & 37.74
& 22.18 & 22.01 & 21.04 & 21.37 & 22.09 \\

& & DARA {\footnotesize \color{blue}{(COLM25)}}
& 27.57 & 33.06 & 34.48 & 37.05 & 36.78
& 13.84 & 15.02 & 17.41 & 20.12 & 22.03 \\

& & MimIC {\footnotesize \color{blue}{(CVPR25)}}
& 36.19 & 36.14 & 38.05 & 37.18 & \underline{37.99}
& 24.11 & 24.02 & 26.07 & 21.91 & 18.02 \\

& & STV {\footnotesize \color{blue}{(AAAI26)}}
& 34.38 & 37.88 & 37.11 & 37.07 & 37.67
& 14.73 & 20.11 & 18.42 & 21.18 & 21.44 \\

\cmidrule(lr){2-2} \cmidrule(lr){3-3}
\cmidrule(lr){4-8} \cmidrule(lr){9-13}

& \multirow{4}{*}{\oursbenchmark}
& MT-ICL {\footnotesize \color{blue}{(ECCVW24)}}
& 32.64 & 33.52 & 33.19 & 32.87 & 32.02
& 24.52 & 24.38 & 23.28 & 23.60 & 24.56 \\

& & DARA {\footnotesize \color{blue}{(COLM25)}}
& 23.83 & 28.37 & 29.88 & 31.12 & 31.47
& 15.68 & 17.26 & 19.90 & 22.65 & 24.17 \\

& & MimIC {\footnotesize \color{blue}{(CVPR25)}}
& 31.41 & 32.08 & 33.42 & 32.66 & 32.37
& \textbf{26.89} & 26.37 & 28.29 & 26.83 & 29.95 \\

& & STV {\footnotesize \color{blue}{(AAAI26)}}
& 29.61 & 32.54 & 32.01 & 31.88 & 32.12
& \textbf{26.89} & 30.87 & \underline{30.59} & \underline{33.46} & \underline{33.90} \\

\cmidrule(lr){1-1} \cmidrule(lr){2-2} \cmidrule(lr){3-3}
\cmidrule(lr){4-8} \cmidrule(lr){9-13}

\multirow{7}{*}{\makecell{Mitigating \\ hallucination}} 
& \multirow{7}{*}{-}
& PAI {\footnotesize \color{blue}{(ECCV24)}}
& 25.18 & 29.62 & 31.08 & 31.43 & 32.07
& 15.48 & 12.67 & 15.26 & 15.84 & 20.72 \\

& & VOLCANO {\footnotesize \color{blue}{(NAACL24)}}
& 32.98 & 33.92 & 34.61 & 35.43 & 34.57
& 22.55 & \underline{30.89} & 23.92 & 23.76 & 24.96 \\

& & LURE {\footnotesize \color{blue}{(ICLR24)}}
& 29.06 & 33.41 & 34.55 & 36.02 & 35.11
& 17.63 & 25.94 & 19.81 & 19.42 & 22.11 \\

& & LRV {\footnotesize \color{blue}{(ICLR24)}}
& 33.05 & 34.47 & 34.92 & 35.11 & 34.02
& 20.74 & 28.96 & 22.41 & 22.18 & 23.07 \\

& & IBD {\footnotesize \color{blue}{(CVPRW25)}}
& 35.44 & 36.83 & 35.27 & 38.02 & 36.88
& 21.83 & 30.12 & 21.95 & 24.11 & 22.64 \\

& & SPARC {\footnotesize \color{blue}{(ICML25)}}
& 34.12 & 36.02 & 37.08 & 36.14 & 36.21
& 19.92 & 27.88 & 20.77 & 21.36 & 23.58 \\

& & \ours (Ours)
& \textbf{38.02} & \textbf{40.71} & \textbf{39.36} & \textbf{41.22} & \textbf{39.18}
& \underline{24.81} & \textbf{32.46} & \textbf{33.92} & \textbf{34.37} & \textbf{35.74} \\

\bottomrule
\end{tabular}
}
\vspace{0.5em}
\caption{
\textbf{Quantitative comparison in multimodal ICL captioning benchmarks.}
For each dataset and shot, the top scores are highlighted in bold, and the second-best scores are underlined.
We use InternVL-3.5-8B as the MLLM and apply random ordering of demonstrations for all baselines except retrieval- and ordering-based methods.
}
\vspace{-0.5em}
\label{tab:icl_captioning_internvl_combined}
\end{table}

\subsection{Details of Baselines}
\label{sec:appx:details_baselines}
Multimodal ICL baselines can be grouped into (\romannumeral 1) retrieval- and ordering-based methods (MMICES~\cite{chen2025can}, VICL~\cite{zhou2024visual}, RICES~\cite{alayrac2022flamingo}), (\romannumeral 2) ICL training-based methods (STV~\cite{ma2026and}, MimIC~\cite{jiang2025mimic}, MT-ICL~\cite{doveh2024towards}, DARA~\cite{chen2025true}), and (\romannumeral 3) task vector-based methods (MTV~\cite{huang2024multimodal}, FV~\cite{todd2024function}).

\subsubsection{Retrieval- and Ordering-based Baselines}

\textbf{MMICES}~\cite{chen2025can} shows that selecting demonstrations based solely on text or visual similarity to the query is suboptimal for in-context learning.
While text similarity is generally more informative than image similarity for selecting ICL demonstrations~\cite{yang2023exploring}, high text similarity often corresponds to generic questions (\eg, ``What does the picture depict?''), which can lead to irrelevant context selection.
Therefore, MMICES considers both modalities: to select $K$ ICL demonstrations, it first retrieves $N$ $(N > K)$ candidates with high visual similarity to the query, and then selects the final $K$ based on text similarity.

\textbf{VICL}~\cite{zhou2024visual} also selects and orders ICL demonstrations considering both visual and text similarity.
Specifically, to select $K$ demonstrations, it first retrieves $N$ $(N > K)$ candidates with high visual similarity to the query image.
It then performs cross-modal reranking to ensure that the selected demonstrations are not only visually similar but also semantically aligned with the query. 
Concretely, it generates a caption $T_q$ describing the query image, and reranks (\ie, ascending order) the $N$ candidates based on cross-modal similarity using CLIP, selecting the top $K$.

\textbf{RICES}~\cite{baldassini2024makes} selects and orders ICL demonstrations based on multimodal similarity to the query. 
Specifically, it computes cosine similarity between the query and candidate samples using embeddings from a multimodal encoder that jointly considers image and text, and selects the most similar samples as demonstrations, ordering them in ascending similarity. 
Following \citet{qin2024factors}, we use BridgeTower~\cite{xu2023bridgetower} as the multimodal encoder.

\subsubsection{ICL training-based Baselines}

\textbf{MT-ICL}~\cite{doveh2024towards} proposes a multi-shot ICL instruction tuning framework for MLLMs. 
Instead of using a single-instruction format (\ie, image + instruction + response), it employs multiple instructions in an ICL format during training, enabling the model to effectively leverage in-context examples. 
In other words, it constructs instruction tuning data in an ICL format and trains MLLMs to utilize demonstrations, thereby explicitly enabling multimodal ICL.

\textbf{DARA}~\cite{chen2025true} shows that MLLMs mainly rely on textual inputs and neglect visual information in the ICL context during multimodal ICL. 
Based on this observation, DARA aims to increase the norm of visual tokens in ICL demonstrations. 
To this end, it trains scaling parameters that are multiplied to the attention norms of visual tokens in the ICL context, thereby enhancing their contribution.
However, since these parameters are trained on ICL datasets, DARA performs well on test sets drawn from the same datasets (\ie, ID datasets) but generalizes poorly to unseen ICL datasets (\ie, OOD datasets), as shown in Fig.~\ref{fig:teaser}.
Moreover, DARA learns shot-specific scaling parameters for $n$-shot ICL, requiring training with $n$-shot data and preventing its use in other shot settings.

\textbf{MimIC}~\cite{jiang2025mimic} argues that prior methods apply shifting vectors to FFN hidden states after attention, whereas such shifts should be applied at the level of individual attention heads. 
Motivated by this, it introduces a modified attention module with learnable parameters that capture the shifting effect. 
Specifically, it replaces standard self-attention heads and trains these parameters to reproduce the outputs obtained when processing ICL demonstrations with the original attention. 
At inference, it applies the learned shift to the query without requiring demonstrations, enabling ICL-like behavior similar to task vector-based methods.

\textbf{STV}~\cite{ma2026and} compute activation deltas between queries with and without demonstrations, and select the top-$k$ heads with the largest difference. 
It then collect activations from these selected heads under ICL and cluster them to form a candidate steering vector set. 
A learnable logit selects which vectors to apply, optimized to minimize the loss between outputs with ICL demonstrations and those with activation steering using selected vectors.
It requires ICL data to construct candidate steering vectors and train the coefficients.

\subsubsection{Task vector-based Baselines}

\textbf{MTV}~\cite{huang2024multimodal} compresses ICL data into task vectors stored in attention-head activations. 
It runs multiple few-shot inferences and averages the last-token attention-head activations to obtain mean task vectors. 
At inference, these vectors steer activations at task-relevant attention-heads that are most useful for the downstream task, allowing the model to exploit many-shot multimodal demonstrations without explicitly placing all examples in the context window.
Similar to STV, it requires ICL data for task vector extraction and task-specific head selections.

\textbf{FV}~\cite{todd2024function} estimates \textit{function vectors}, which trigger specific functions (\eg, translation or synonym generation) when applied to the transformer hidden states.
Specifically, it identifies attention heads whose output probabilities are most sensitive to label shuffling in ICL demonstrations, and defines a task vector by summing their activations over an ICL dataset. 
During inference, this task vector is added to the hidden states for a query, enabling in-context learning behavior even without explicit demonstrations and alleviating context length constraints.

\subsection{Details of Prompt Decomposition in \ours}
\label{sec:appx:prompt_decomposition}
In \ours, we decompose captions generated by the MLLM into proposition-level components using GPT-4o-mini. 
During decomposition, we use the following prompt:

\begin{tcolorbox}[width=\linewidth, colframe=black, before skip=2mm, after skip=0.0cm, boxsep=0.0cm, middle=0.0cm, top=0.3cm, bottom=0.3cm]
\small
\ttfamily
You are given two sentences: Sentence 1 and Sentence 2. Decompose them into three full-sentence propositions. \\ \\

Rules: \\ 

1. Output exactly three fields: \\
   - shared \\
   - variant 1 \\ 
   - variant 2 \\

2. shared: \\ 
   - must include only the meaning that is common to both Sentence 1 and Sentence 2 \\ 
   - must not include content that differs between the two sentences \\

3. variant 1: \\
   - must include only the meaning that is unique to Sentence 1 \\
   - must be a full sentence \\
   - must not repeat the shared content unless needed for grammaticality \\

4. variant 2: \\
   - must include only the meaning that is unique to Sentence 2 \\
   - must be a full sentence \\
   - must not repeat the shared content unless needed for grammaticality \\

5. Preserve the original meaning of both sentences. \\ \\
6. Do not add extra facts. \\
\end{tcolorbox}

\subsection{Details of Multimodal Contrastive Training Pairs}
\label{sec:appx:cont_datasets}

\begin{figure}[t]
    \centering   
    \includegraphics[width=\linewidth]{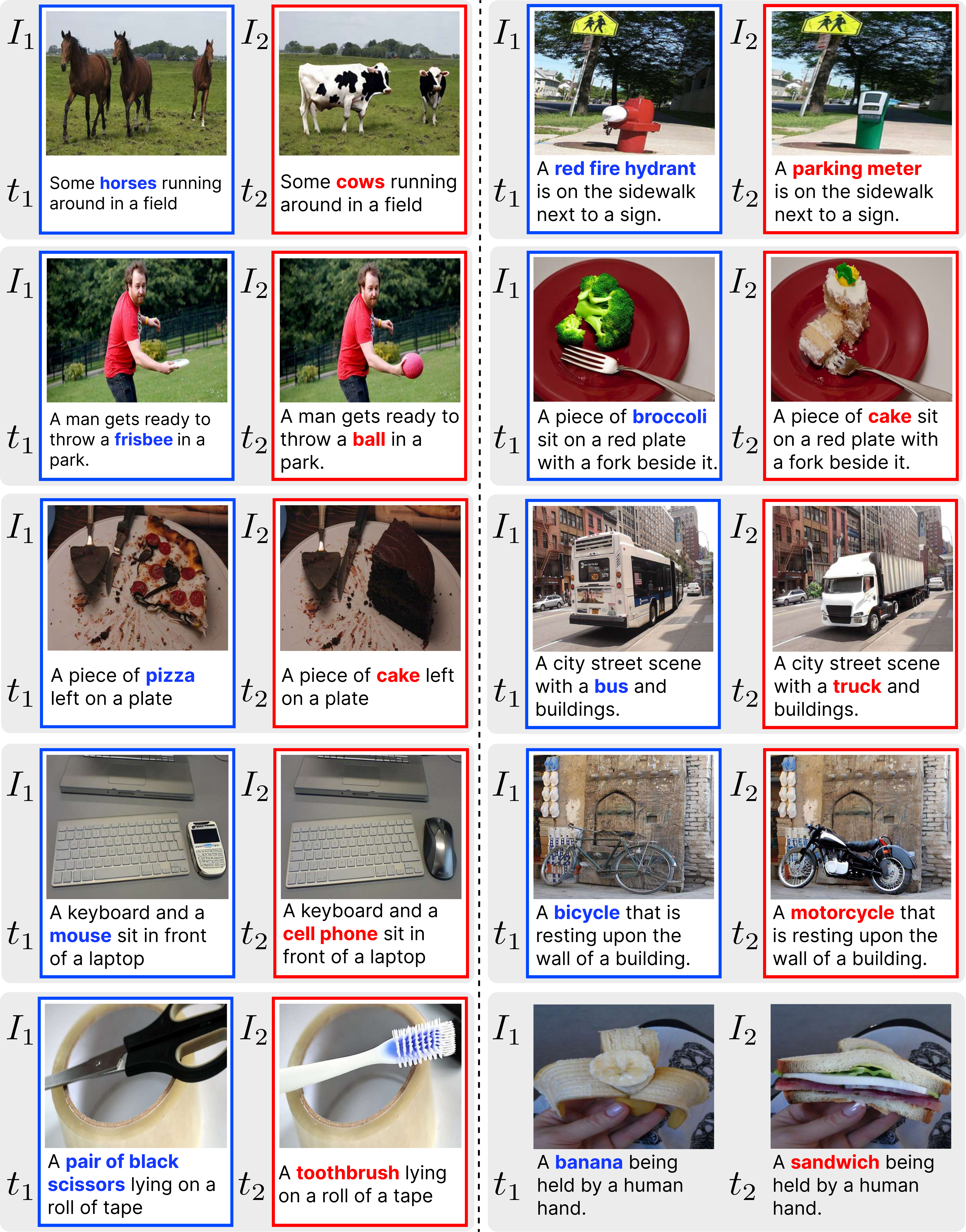}
    \vspace{-1em}
    \caption{
        \textbf{Examples of multimodal contrastive datasets.}
        Each sample consists of two image-text pairs, $(I_1, t_1)$ and $(I_2, t_2)$, with captions $t_1$ and $t_2$ describe images $I_1$ and $I_2$, respectively.
        $I_1$ and $I_2$ are visually contrastive (\eg, differing in objects (left) or object attributes (right)).
    }
    \label{fig:contrastive_data_examples}
\end{figure}

We first illustrate examples of multimodal contrastive training pairs in Fig.~\ref{fig:contrastive_data_examples}
We also argue that our method generalizes to any multimodal contrastive dataset with such image–caption pairs. 
In Tab.~\ref{tab:contrastive_benchmark_comparison}, we report the performance of \ours across diverse multimodal contrastive benchmarks (InpaintCOCO~\cite{rosch2024enhancing}, VisMin~\cite{awal2024vismin}, Winoground~\cite{thrush2022winoground}, and EQBEN~\cite{wang2023equivariant}) when trained on each of them.
As shown, \ours achieves consistent performance regardless of the contrastive training benchmark, indicating that gains stem from the proposed contrastive reward rather than any specific dataset. 
To better capture diverse visual contrastive characteristics, we construct a mixed training set by randomly sampling 500 pairs each from EQBEN and InpaintCOCO, which exhibit the most distinct and complementary properties.
Specifically, EQBEN primarily varies object behavior and count, whereas InpaintCOCO focuses more on differences in object identity, attributes, and relations.
We expect that combining \ours with more diverse, high-quality multimodal contrastive datasets will further enhance visual grounding.

\subsection{Quantitative Analysis across Model Sizes}
\label{sec:appx:diverse_sizes}
We employ Qwen-VL-3-8B model main experiments, as well as InternVL-3.5-8B and LLaVA-OneVision-1.5-8B in Sec.~\ref{sec:appx:diverse_models}.
To validate that \ours improves multimodal ICL performance regardless of the base MLLM size, we further conduct experiments with 4B and 14B models.
Specifically, we include Qwen-VL-3-4B and InternVL-3.5-14B.
Since Qwen-VL-3 is only available up to the 8B scale, we use the 14B variant of InternVL-3.5 to evaluate the effectiveness of \ours on larger models.
As shown in Tab.~\ref{tab:icl_captioning_qwen-4b} and Tab.~\ref{tab:icl_captioning_internvl14b}, \ours consistently outperforms baselines, which aligns with the results observed on 8B models, demonstrating that its effectiveness is not limited to a specific model scale.

\begin{table}[t!]
\centering
\small
\resizebox{\linewidth}{!}{
\begin{tabular}{lclcccccccccc}
\toprule
\multirow{2}{*}{\textbf{Category}} & \multirow{2}{*}{\makecell{\textbf{ICL Training}\\ \textbf{Dataset}}} & \multirow{2}{*}{\textbf{Method}} & \multicolumn{5}{c}{\textbf{MS COCO}} & \multicolumn{5}{c}{\textbf{PICL}} \\ 
& & & 0-shot & 1-shot & 2-shot & 4-shot & 8-shot & 0-shot & 1-shot & 2-shot & 3-shot & 4-shot \\

\cmidrule(lr){1-1} \cmidrule(lr){2-2} \cmidrule(lr){3-3} \cmidrule(lr){4-8} \cmidrule(lr){9-13}

Off-the-shelf & - 
& -
& 11.02 & 14.87 & 16.55 & 21.11 & 22.34
& 22.31 & 26.94 & 22.17 & 30.88 & 31.42 \\

\cmidrule(lr){1-1} \cmidrule(lr){2-2} \cmidrule(lr){3-3} \cmidrule(lr){4-8} \cmidrule(lr){9-13}

\multirow{3}{*}{\makecell{Retrieval and \\ ordering}} & -
& RICES {\footnotesize \color{blue}{(NeurIPS22)}}
& 13.21 & 17.44 & 20.02 & 21.08 & 24.91
& 26.77 & 24.01 & 27.83 & 25.16 & 29.94 \\

& & VICL {\footnotesize \color{blue}{(ACL24)}}
& 12.88 & 19.13 & 20.76 & 22.05 & 24.77
& 20.42 & 29.63 & 28.91 & 28.04 & 32.11 \\

& & MMICES {\footnotesize \color{blue}{(WACV25)}}
& 10.74 & 15.62 & 18.03 & 20.14 & 23.09
& 21.55 & 27.48 & 18.92 & 25.77 & 27.61 \\

\cmidrule(lr){1-1} \cmidrule(lr){2-2} \cmidrule(lr){3-3} \cmidrule(lr){4-8} \cmidrule(lr){9-13}

\multirow{4}{*}{Task vector} & \multirow{2}{*}{MS COCO}
& MTV {\footnotesize \color{blue}{(NeurIPS24)}}
& 6.11 & 10.84 & 14.73 & 17.02 & 20.88
& 24.41 & 22.88 & 18.72 & 28.11 & 30.92 \\

& & FV {\footnotesize \color{blue}{(ICLR24)}}
& 9.87 & 13.55 & 17.12 & 20.44 & 23.02
& 23.88 & 22.11 & 26.73 & 29.04 & 30.55 \\

\cmidrule(lr){2-2} \cmidrule(lr){3-3} \cmidrule(lr){4-8} \cmidrule(lr){9-13}

& \multirow{2}{*}{PICL}
& MTV {\footnotesize \color{blue}{(NeurIPS24)}}
& 2.84 & 7.11 & 11.02 & 13.74 & 16.42
& 22.18 & 24.91 & 20.11 & 26.03 & 28.72 \\

& & FV {\footnotesize \color{blue}{(ICLR24)}}
& 5.98 & 10.44 & 13.37 & 16.09 & 19.11
& 25.92 & 27.83 & 23.74 & 29.62 & 32.41 \\

\cmidrule(lr){1-1} \cmidrule(lr){2-2} \cmidrule(lr){3-3} \cmidrule(lr){4-8} \cmidrule(lr){9-13}

\multirow{8}{*}{ICL training} & \multirow{4}{*}{MS COCO}
& MT-ICL {\footnotesize \color{blue}{(ECCVW24)}}
& 31.85 & 35.92 & 36.14 & 37.21 & \textbf{39.66}
& \textbf{31.66} & 27.04 & 33.42 & 28.71 & 34.87 \\

& & DARA {\footnotesize \color{blue}{(COLM25)}}
& 18.54 & 23.61 & 25.74 & 28.33 & 31.88
& 20.11 & 21.83 & 26.34 & 24.01 & 29.41 \\

& & MimIC {\footnotesize \color{blue}{(CVPR25)}}
& 22.07 & 28.19 & 29.44 & 30.01 & 33.02
& 23.64 & 28.94 & 26.77 & 29.91 & 30.72 \\

& & STV {\footnotesize \color{blue}{(AAAI26)}}
& 31.44 & 36.28 & 35.11 & 36.52 & 37.02
& 31.88 & 33.27 & 29.44 & 35.02 & 34.11 \\

\cmidrule(lr){2-2} \cmidrule(lr){3-3} \cmidrule(lr){4-8} \cmidrule(lr){9-13}

& \multirow{4}{*}{PICL}
& MT-ICL {\footnotesize \color{blue}{(ECCVW24)}}
& 28.03 & 33.12 & 33.77 & 34.88 & 35.14
& 29.84 & 35.27 & 30.91 & 36.88 & 33.02 \\

& & DARA {\footnotesize \color{blue}{(COLM25)}}
& 14.77 & 20.05 & 21.33 & 24.61 & 28.22
& 22.44 & 28.13 & 24.28 & 30.55 & 31.92 \\

& & MimIC {\footnotesize \color{blue}{(CVPR25)}}
& 19.31 & 25.77 & 26.02 & 27.88 & 29.65
& 25.31 & 31.44 & 24.88 & 29.77 & 33.65 \\

& & STV {\footnotesize \color{blue}{(AAAI26)}}
& 27.66 & 32.44 & 33.18 & 34.05 & 35.27
& 30.22 & 35.91 & 31.08 & 33.44 & 37.11 \\

\cmidrule(lr){1-1} \cmidrule(lr){2-2} \cmidrule(lr){3-3} \cmidrule(lr){4-8} \cmidrule(lr){9-13}

\multirow{7}{*}{\makecell{Mitigating \\ hallucination}} & -
& PAI {\footnotesize \color{blue}{(ECCV24)}}
& 8.12 & 13.44 & 14.02 & 18.76 & 20.01
& 18.88 & 27.02 & 24.76 & 22.48 & 29.91 \\

& & VOLCANO {\footnotesize \color{blue}{(NAACL24)}}
& 24.88 & 30.22 & 28.44 & 32.03 & 36.55
& 25.55 & 29.84 & 22.13 & 29.02 & 31.66 \\

& & LURE {\footnotesize \color{blue}{(ICLR24)}}
& 10.23 & 12.11 & 15.04 & 18.33 & 20.52
& 23.77 & 24.63 & 27.91 & 31.18 & 27.94 \\

& & LRV {\footnotesize \color{blue}{(ICLR24)}}
& 12.77 & 22.94 & 21.33 & 19.01 & 24.12
& \underline{28.91} & 28.55 & 31.72 & 32.87 & 35.44 \\

& & IBD {\footnotesize \color{blue}{(CVPRW25)}}
& 25.44 & 28.01 & 30.88 & 31.55 & 29.03
& 17.91 & 24.84 & 24.72 & 21.03 & 23.98 \\

& & SPARC {\footnotesize \color{blue}{(ICML25)}}
& 27.11 & 30.42 & 29.18 & 34.02 & 36.77
& 26.42 & 33.14 & 28.11 & 30.52 & 35.62 \\

& & \ours (Ours)
& \textbf{32.88} & \textbf{39.12} & \textbf{38.77} & \textbf{39.55} & \underline{38.21}
& 27.55 & \textbf{38.72} & \textbf{37.41} & \textbf{39.84} & \textbf{35.93} \\

\bottomrule
\end{tabular}
}
\vspace{0.5em}
\caption{
\textbf{Quantitative comparison in multimodal ICL captioning benchmarks.}
For each dataset and shot, the top scores are highlighted in bold, and the second-best scores are underlined.
We use Qwen-VL-3-4B as the MLLM and apply random ordering of demonstrations for all baselines except retrieval- and ordering-based methods.
}
\vspace{-1em}
\label{tab:icl_captioning_qwen-4b}
\end{table}

\begin{table}[t!]
\centering
\small
\resizebox{\linewidth}{!}{
\begin{tabular}{lclcccccccccc}
\toprule
\multirow{2}{*}{\textbf{Category}} 
& \multirow{2}{*}{\makecell{\textbf{ICL Training}\\ \textbf{Dataset}}} 
& \multirow{2}{*}{\textbf{Method}} 
& \multicolumn{5}{c}{\textbf{MS COCO}} 
& \multicolumn{5}{c}{\textbf{\oursbenchmark}} \\ 
& & 
& 0-shot & 1-shot & 2-shot & 4-shot & 8-shot 
& 0-shot & 1-shot & 2-shot & 3-shot & 4-shot \\

\cmidrule(lr){1-1} \cmidrule(lr){2-2} \cmidrule(lr){3-3}
\cmidrule(lr){4-8} \cmidrule(lr){9-13}

Off-the-shelf & - & Random
& 26.41 & 35.02 & 35.11 & 32.84 & 35.91
& 24.82 & 17.77 & 28.66 & 22.48 & 29.14 \\

\cmidrule(lr){1-1} \cmidrule(lr){2-2} \cmidrule(lr){3-3}
\cmidrule(lr){4-8} \cmidrule(lr){9-13}

\multirow{3}{*}{\makecell{Retrieval and \\ ordering}} 
& \multirow{3}{*}{-}
& RICES {\footnotesize \color{blue}{(NeurIPS22)}}
& 24.91 & 32.66 & 36.58 & 34.11 & 31.94
& 26.04 & 28.85 & 25.71 & 34.72 & 30.63 \\

& & VICL {\footnotesize \color{blue}{(ACL24)}}
& 23.84 & 36.02 & 36.44 & 35.97 & 34.82
& 22.51 & 19.47 & 27.18 & 31.11 & 33.62 \\

& & MMICES {\footnotesize \color{blue}{(WACV25)}}
& 28.73 & 30.42 & 34.95 & 36.48 & 37.28
& 19.18 & 15.77 & 21.42 & 29.83 & 34.66 \\

\cmidrule(lr){1-1} \cmidrule(lr){2-2} \cmidrule(lr){3-3}
\cmidrule(lr){4-8} \cmidrule(lr){9-13}

\multirow{4}{*}{Task vector} 
& \multirow{2}{*}{MS COCO}
& MTV {\footnotesize \color{blue}{(NeurIPS24)}}
& 22.74 & 34.18 & 29.41 & 33.02 & 35.88
& 21.92 & 18.74 & 30.11 & 23.84 & 31.27 \\

& & FV {\footnotesize \color{blue}{(ICLR24)}}
& 30.01 & 36.47 & 37.88 & 31.77 & 36.41
& 27.72 & 24.83 & 28.96 & 26.15 & 32.77 \\

\cmidrule(lr){2-2} \cmidrule(lr){3-3}
\cmidrule(lr){4-8} \cmidrule(lr){9-13}

& \multirow{2}{*}{\oursbenchmark}
& MTV {\footnotesize \color{blue}{(NeurIPS24)}}
& 20.11 & 30.74 & 27.02 & 29.88 & 31.41
& 23.18 & 16.21 & 28.74 & 25.92 & 29.03 \\

& & FV {\footnotesize \color{blue}{(ICLR24)}}
& 24.63 & 34.19 & 30.55 & 33.26 & 35.94
& 27.64 & 20.55 & 32.41 & 29.84 & 33.76 \\

\cmidrule(lr){1-1} \cmidrule(lr){2-2} \cmidrule(lr){3-3}
\cmidrule(lr){4-8} \cmidrule(lr){9-13}

\multirow{8}{*}{ICL training} 
& \multirow{4}{*}{MS COCO}
& MT-ICL {\footnotesize \color{blue}{(ECCVW24)}}
& \underline{37.91} & \underline{41.22} & 34.88 & \underline{41.03} & 35.66
& 28.11 & 28.84 & 33.96 & 28.74 & 33.88 \\

& & DARA {\footnotesize \color{blue}{(COLM25)}}
& 31.94 & 32.61 & 35.27 & 40.72 & 37.83
& 17.11 & 25.76 & 19.22 & 32.47 & 25.88 \\

& & MimIC {\footnotesize \color{blue}{(CVPR25)}}
& 39.55 & 36.83 & \underline{41.28} & 39.47 & \underline{41.51}
& 36.88 & 30.91 & 33.77 & 32.13 & 29.06 \\

& & STV {\footnotesize \color{blue}{(AAAI26)}}
& 31.74 & 35.08 & 37.44 & 40.61 & 36.02
& 26.73 & 28.12 & 20.91 & 33.76 & 28.55 \\

\cmidrule(lr){2-2} \cmidrule(lr){3-3}
\cmidrule(lr){4-8} \cmidrule(lr){9-13}

& \multirow{4}{*}{\oursbenchmark}
& MT-ICL {\footnotesize \color{blue}{(ECCVW24)}}
& 30.02 & 34.62 & 36.11 & 30.14 & 35.48
& 29.22 & 25.18 & 30.74 & 27.84 & 37.61 \\

& & DARA {\footnotesize \color{blue}{(COLM25)}}
& 24.44 & 30.98 & 26.84 & 32.58 & 29.41
& 25.61 & 21.72 & 32.94 & 35.55 & 26.94 \\

& & MimIC {\footnotesize \color{blue}{(CVPR25)}}
& 28.11 & 34.55 & 36.74 & 30.62 & 34.83
& \textbf{38.66} & 35.03 & 30.71 & 40.66 & 37.94 \\

& & STV {\footnotesize \color{blue}{(AAAI26)}}
& 32.44 & 35.22 & 31.55 & 34.91 & 35.76
& 30.55 & 39.83 & \underline{38.61} & \underline{44.42} & \underline{38.77} \\

\cmidrule(lr){1-1} \cmidrule(lr){2-2} \cmidrule(lr){3-3}
\cmidrule(lr){4-8} \cmidrule(lr){9-13}

\multirow{7}{*}{\makecell{Mitigating \\ hallucination}} 
& \multirow{7}{*}{-}
& PAI {\footnotesize \color{blue}{(ECCV24)}}
& 23.74 & 26.55 & 29.02 & 33.84 & 31.12
& 18.14 & 25.94 & 27.11 & 29.88 & 24.02 \\

& & VOLCANO {\footnotesize \color{blue}{(NAACL24)}}
& 35.12 & 34.84 & 36.52 & 31.98 & 37.81
& 30.55 & 41.88 & 32.31 & 38.62 & 33.76 \\

& & LURE {\footnotesize \color{blue}{(ICLR24)}}
& 28.81 & 31.27 & 36.84 & 38.12 & 32.66
& 30.83 & 28.91 & 29.74 & 24.02 & 25.18 \\

& & LRV {\footnotesize \color{blue}{(ICLR24)}}
& 36.44 & 32.11 & 38.52 & 36.73 & 31.62
& 28.77 & 37.03 & 33.91 & 36.72 & 26.18 \\

& & IBD {\footnotesize \color{blue}{(CVPRW25)}}
& 39.27 & 37.66 & 31.88 & 39.94 & 40.21
& 29.11 & \underline{42.74} & 32.63 & 36.83 & 28.41 \\

& & SPARC {\footnotesize \color{blue}{(ICML25)}}
& 35.94 & 39.88 & 40.44 & 35.12 & 39.62
& 32.77 & 39.66 & 24.61 & 33.88 & 34.92 \\

& & \ours (Ours)
& \textbf{40.88} & \textbf{43.61} & \textbf{42.84} & \textbf{45.33} & \textbf{41.77}
& \underline{36.62} & \textbf{43.91} & \textbf{40.88} & \textbf{45.27} & \textbf{40.42} \\

\bottomrule
\end{tabular}
}
\vspace{0.5em}
\caption{
\textbf{Quantitative comparison in multimodal ICL captioning benchmarks.}
For each dataset and shot, the top scores are highlighted in bold, and the second-best scores are underlined.
We use InternVL-3.5-14B as the MLLM and apply random ordering of demonstrations for all baselines except retrieval- and ordering-based methods.
}
\label{tab:icl_captioning_internvl14b}
\end{table}

\subsection{Extended Related Work}

\paragraph{Multimodal In-Context Learning.}
Building on the emergence of ICL in LLMs, prior work has attempted to extend ICL to MLLMs, which can be broadly categorized into three groups: (i) retrieval- and ordering-based methods, (ii) ICL training-based methods, and (iii) task vector-based methods.

Retrieval- and ordering-based methods~\cite{baldassini2024makes, luo2024does, chen2025can} retrieve and rank ICL demonstrations by their similarity to the query, improving performance with appropriate sample selection and ordering.
However, they are highly sensitive to prompt design and exhibit substantial performance fluctuations~\cite{guo2025deepseek}, especially in multimodal settings, where aligning textual and visual features introduces additional complexity~\cite{lu2022fantastically, qin2024factors, li2025m}.
As a result, they often perform worse than random selection~\cite{yang2024lever}.

ICL training-based methods~\cite{doveh2024towards, tai2024link, chen2025true, li2025m, jiang2025mimic} enable multimodal ICL by further training MLLMs on ICL datasets, either over full model parameters~\cite{doveh2024towards, tai2024link} or additional learnable modules~\cite{chen2025true}. 
While these approaches achieve strong performance on the same ICL datasets used for training, they show limited generalization, with significant performance drops on ICL benchmarks not seen during training.
Moreover, additional training with ICL data can lead to forgetting of previously well-performing tasks (\eg, simple VQA tasks), as shown in Fig.~\ref{fig:forgetting}.

Task vector-based methods~\cite{peng2024live, liu2024context, huang2024multimodal, ma2026and} compress ICL demonstrations into compact task vectors that steer model activations during inference, achieving effects comparable to conditioning on explicit demonstrations.
While they reduce memory overhead and context length constraints and perform well on simple tasks, they struggle to generalize to complex multimodal scenarios~\cite{peng2024live, li2025m} and may lose task-specific details due to limited representational richness~\cite{huang2024multimodal, ma2026and}. 
Furthermore, similar to ICL training-based methods, they show limited generalization to unseen ICL benchmarks (Fig.~\ref{fig:teaser}). 
Different from prior work that improves multimodal ICL through better demonstration selection or ICL data-dependent adaptation, our method targets a more fundamental bottleneck, \ie, poor visual grounding, improving multimodal ICL without relying on any ICL datasets.

\paragraph{Mitigating Hallucinations in MLLMs.}
\label{sec:appx:related_work}
Instruction tuning data enhancement methods~\cite{liu2024mitigating, wang2024vigc, yu2024hallucidoctor} mitigate bias and hallucinations in visual instruction tuning data, improving factual grounding.
\citet{zhu2024minigpt} show that these data are biased toward positive responses, while \citet{liu2024mitigating, yu2024hallucidoctor} report that over 30\% of LLaVA-Instruct-150K contains hallucinated instructions.
To address this, prior work performs additional instruction tuning on remedial datasets~\cite{lu2023evaluation, zhu2024minigpt} (\eg, with balanced positive and negative responses) or preprocesses instruction tuning data to detect and correct hallucination~\cite{yu2024hallucidoctor, wang2024vigc}.
However, these approaches require access to instruction tuning data and incur substantial computational cost due to their large scale.

Supplementary network-based approaches~\cite{zhou2024analyzing, wang2024mitigating, heiman2025factchexcker} use auxilary networks to identify hallucination of the output or to revise the output text from LVLMs. 
LURE~\cite{zhou2024analyzing} trains a reviser network on hallucinated–clean caption pairs to refine MLLM outputs, while FactChecker~\cite{heiman2025factchexcker} generates report-specific code to verify facts in each medical report and update them accordingly.
However, these approaches typically require a substantial number of additional parameters or training data~\cite{zhu2025ibd}.

Visual grounding enhancement approaches often directly amplify attention to visual tokens during self-attention to counter the strong bias of MLLMs toward textual context over visual inputs~\cite{liu2024paying, gao2025aim, chen2025true}.
PAI~\cite{liu2024paying} amplifies all visual token attention by multiplying with a model-specific scaling factor, but this indiscriminate scaling can increase attention to irrelevant tokens and exacerbate hallucinations~\cite{jung2025visual}.
SPARC~\cite{jung2025visual} instead selectively amplifies only the attention of relevant visual tokens through important token selection.
However, directly scaling visual token norms can distort feature representations, as it alters the magnitude of embeddings~\cite{zeng2021improving}. 

\begin{figure}[t!]
    \centering   
    \includegraphics[width=\linewidth]{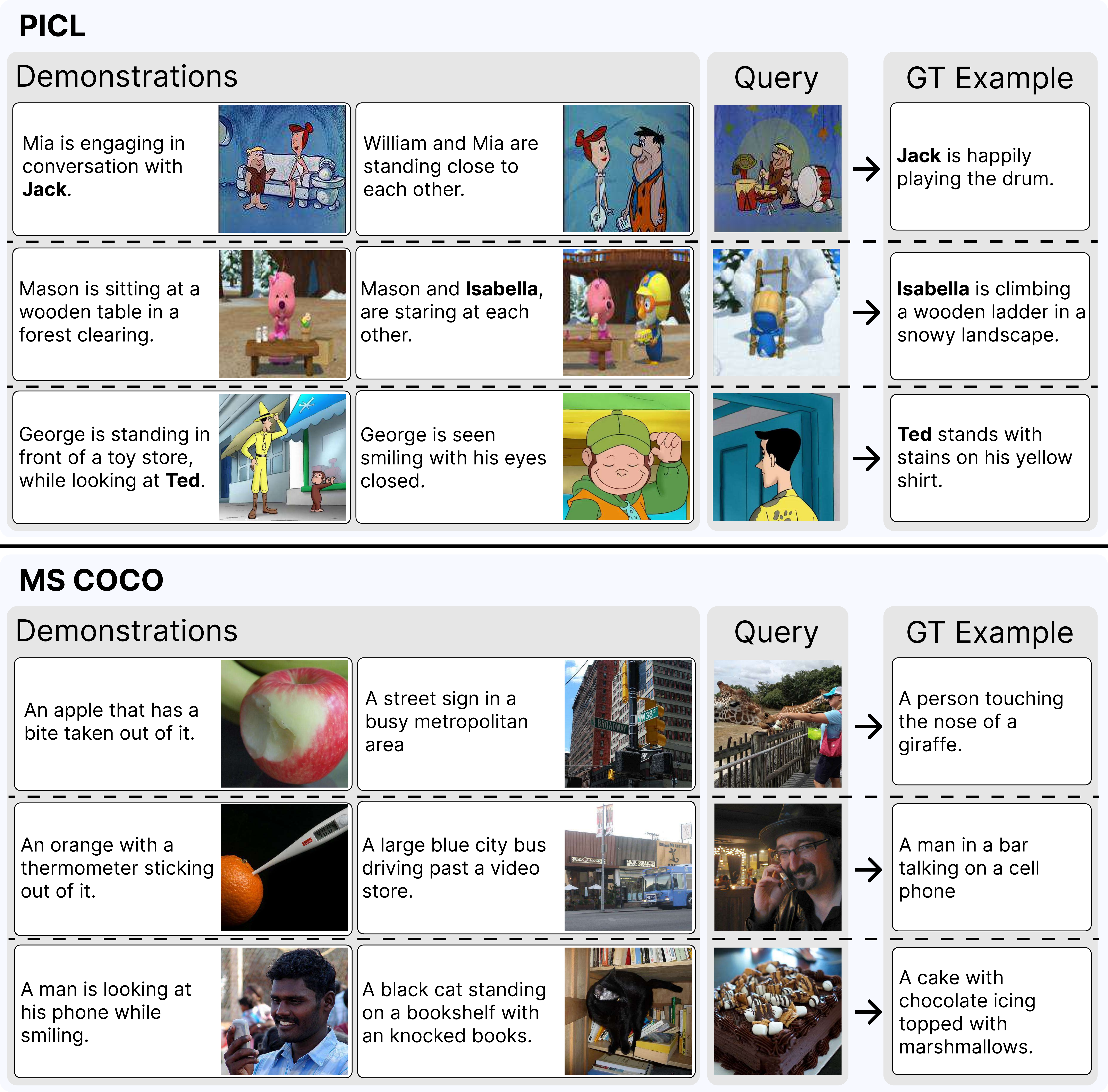}
    \vspace{-1em}
    \caption{
        \textbf{Examples comparing \oursbenchmark and MS COCO.}
        While \oursbenchmark requires visual understanding all demonstrations to answer correctly (\eg, resolving ``Mason'' in the first demo is necessary to correctly identify ``Isabella'' for Pororo in the second, enabling the query prediction), MS COCO ICL task can often be solved using pre-trained priors without understanding demonstrations.
    }
    \label{fig:additional_examples_picl}
\end{figure}

\subsection{Additional Examples of \oursbenchmark}
\label{sec:appx:benchmark_examples}

In addition to Fig.~\ref{fig:additional_examples_picl}, we further provide examples of \oursbenchmark by comparing it with MS COCO~\cite{lin2014microsoft}, existing multimodal ICL captioning benchmark.
As shown, MS COCO ICL mainly involves generic captioning tasks, where query captions can often be predicted using pre-trained semantic priors without truly understanding the demonstrations, although the demonstrations may still guide the output format.

In contrast, \oursbenchmark requires visual understanding across all demonstrations to answer the query. 
For example, in the first row of Fig.~\ref{fig:additional_examples_picl}, the first demonstration alone is insufficient to identify which character is `Jack' or `Mia'. 
However, the second demonstration reintroduces only the female character and mentions `Mia', allowing the model to infer that the male character in the first demonstration is `Jack'—knowledge necessary to correctly answer the query as `Jack'.
Similarly, in the second and third rows, accurate visual understanding of each demonstration is required to produce the correct answer.
Consequently, \oursbenchmark evaluate true visual grounding in multimodal ICL and preventing shortcut solutions based solely on pre-trained semantic priors.

\begin{table*}[t]
\vspace{-0.3em}
\centering
\resizebox{\linewidth}{!}{
\begin{tabular}{lcccccccccc}
\toprule
\multirow{2}{*}{\textbf{Contrastive Benchmark}} 
& \multicolumn{5}{c}{\textbf{TrueMICL}} 
& \multicolumn{5}{c}{\textbf{MS COCO}} \\

& 0-shot & 1-shot & 2-shot & 4-shot & 8-shot
& 0-shot & 1-shot & 2-shot & 4-shot & 8-shot \\

\cmidrule(lr){1-1}
\cmidrule(lr){2-6}
\cmidrule(lr){7-11}

Winoground
& 28.74 & 48.12 & 52.03 & \textbf{64.91} & 66.77
& 36.84 & 41.72 & 43.03 & 42.11 & 43.88 \\

VisMin 
& 28.74 & 49.72 & 54.18 & 60.05 & 65.11 
& \textbf{40.07} & 44.82 & 44.44 & 45.11 & 43.96 \\

InpaintCOCO 
& \textbf{31.11} & 49.63 & \textbf{54.62} & 63.80 & \textbf{68.24}
& 38.21 & 45.02 & \textbf{45.18} & 43.77 & \textbf{46.91} \\

EqBen 
& 27.41 & \textbf{52.88} & 53.03 & 59.44 & 63.02 
& 38.72 & \textbf{47.03} & 41.92 & \textbf{48.21} & 41.33 \\






\bottomrule
\end{tabular}
}
\vspace{-0.3em}
\caption{
    \textbf{Quantitaive comparison of \ours across different contrastive benchmarks.}
    For each dataset and shot, the top scores are highlighted in bold, and the second-best scores are underlined.
    For a fair comparison, we randomly sample 1,000 contrastive pairs from each benchmark.
    We use Qwen-VL-3-8B as the base MLLM for all contrastive benchmarks.
    }
\label{tab:contrastive_benchmark_comparison}
\vspace{-.6em}
\end{table*}

\subsection{Details of \oursbenchmark}
\label{sec:appx:our_benchmark_details}

\subsubsection{Details of Benchmark Construction}
\oursbenchmark is constructed from visual storytelling datasets where the same characters appear across multiple frames, each with a corresponding caption, including PororoSV~\cite{li2019storygan}, FlintstonesSV~\cite{gupta2018imagine}, and StoryStream~\cite{yang2025seed}.
The construction consists of two steps: (i) data preprocessing and (ii) ICL data construction.


\paragraph{Data Preprocessing.}
Each dataset contains a finite set of characters $\mathcal{C}$ (\eg, $\mathcal{C}=\{\text{Pororo, Loopy, Crong, Poby, Eddy}\}$ in PororoSV).
We define a reference image for each character in $\mathcal{C}$. 
Then, using SigLIP~\cite{zhai2023sigmoid} and the reference images, we determine which characters are present in each frame and group frames according to the set of characters present in each frame.

We then rewrite the corresponding caption for each frame using GPT-4o-mini. Although visual storytelling datasets provide captions for each frame, these are often misaligned with the frame or inaccurate, as they are derived from video-level descriptions and may not precisely reflect frame-level content. 
However, frame-only captioning cannot reliably identify character names, even when the set of present characters is known, as it cannot determine which name corresponds to which entity when multiple characters appear.
To prevent this, we provide the reference images $\mathcal{C}$ as context and rewrite the captions accordingly, using the following prompt:

\begin{tcolorbox}[width=\linewidth, colframe=black, before skip=2mm, after skip=0.0cm, boxsep=0.0cm, middle=0.0cm, top=0.3cm, bottom=0.3cm]
\small
\ttfamily

You are given an image and a set of reference images. Write a caption describing the image. \\ \\

Rules: \\
- Generate a caption for the image using the reference images to identify characters \\
- Add visual information based only on the image \\
- Focus on visible objects, interactions, positions, and background elements \\
- Prefer concrete scene details over adjectives or adverbs \\
- Do not invent unsupported details \\
- Do not make assumptions \\
\end{tcolorbox}
\vspace{1em}

Next, we replace original (\ie, common) identifiers with personalized ones. Specifically, we generate English names using GPT-4o-mini~\cite{Achiam2023GPT4TR} and substitute each identifier with a randomly selected name.
To generate English names, we use following prompt:

\begin{tcolorbox}[width=\linewidth, colframe=black, before skip=2mm, after skip=0.0cm, boxsep=0.0cm, middle=0.0cm, top=0.3cm, bottom=0.3cm]
\small
\ttfamily

Generate 100 general English first names. \\ \\

Rules: \\
- Include a balanced mix of male and female names. \\
- Use common, natural-sounding English names. \\
- Avoid rare, fantasy, historical, or celebrity-specific names. \\
- Each name should be a single first name only. \\
- No duplicate names.
\end{tcolorbox}
\vspace{1em}

\begin{figure}[t!]
    \centering   
    \includegraphics[width=\linewidth]{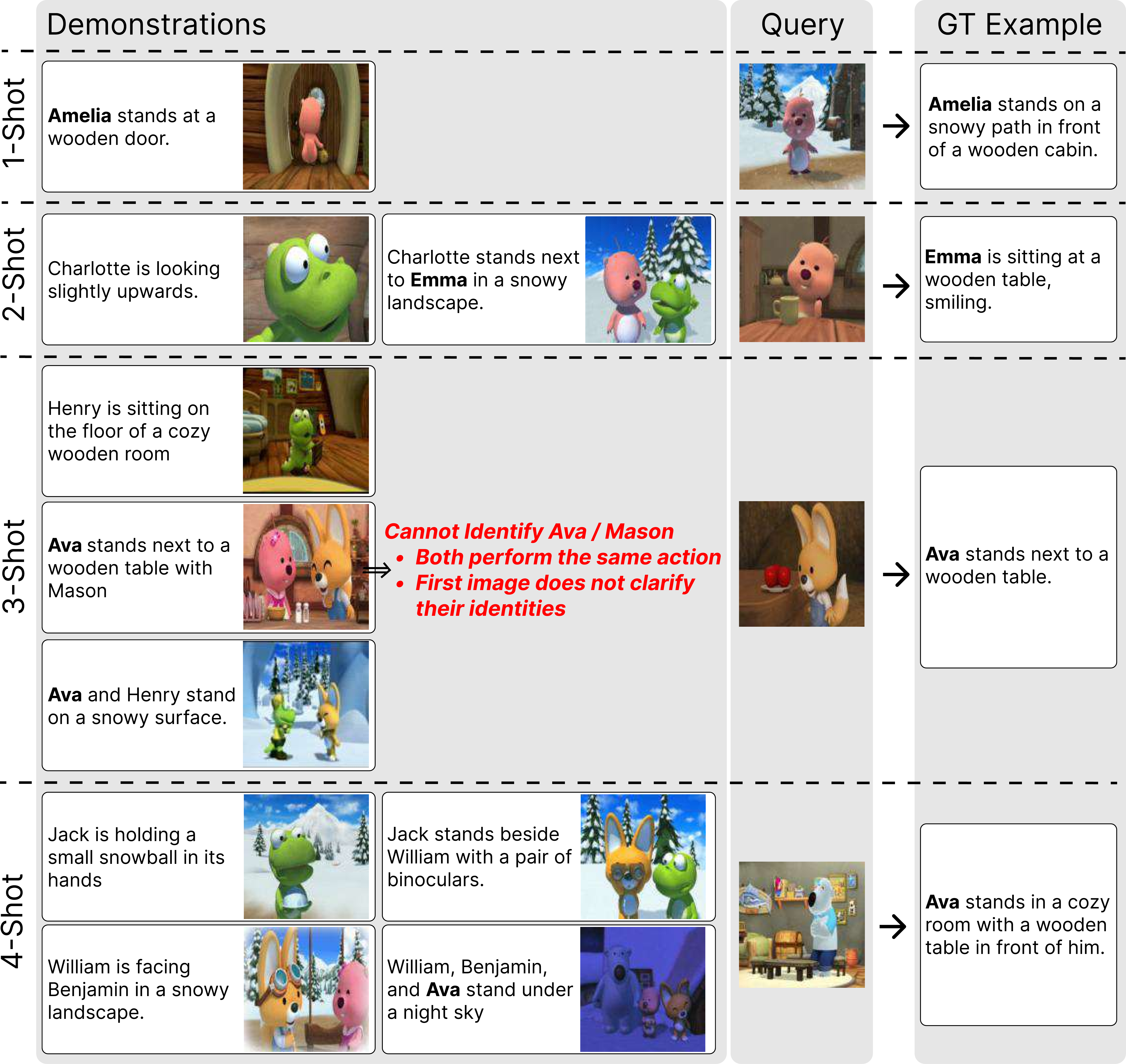}
    \vspace{-1em}
    \caption{
        \textbf{Comparison across ICL-shot settings.}
        Higher-shot settings require reasoning over increasingly complex relationships among demonstrations, rather than simply providing more context.
        For example, in the 3-shot case, the first demo lets the model infer that the dinosaur is ``Henry'', while the second remains ambiguous since ``Ava'' and ``Mason'' perform the same action; only after the third demo (where the fox is identified as ``Henry'') can the model resolve the ambiguity and infer that the pink character in the second demo is ``Ava''.
        In contrast, lower-shot settings can be easier due to fewer required inferences (e.g., in 1-shot, the pink character can be directly identified as ``Amella'').
        }
    \label{fig:picl_shot}
\end{figure} 

\paragraph{ICL Data Construction.}
After preprocessing, we manually select a subset of frames (\ie, image–personalized caption pairs) from each sample in the visual storytelling datasets to construct ICL demonstrations and queries.
Specifically, frames are selected based on the following criteria:
(i) characters in the query image appear in at least one demonstration;
(ii) demonstrations include at least two distinct characters;
(iii) demonstrations provide sufficient context to distinguish character identities.

The first criterion is a necessary condition for solving personalized ICL, as demonstrations provide the only cues for identifying the correct identifier in the query image.
The second criterion ensures evaluation of true visual grounding. 
If all demonstrations contain only a single personalized entity, the model may predict query caption by reusing the subject from the text demonstrations without understanding visual evidence. 
In such cases, it is unclear whether the correct identifier is derived from visual grounding or merely copied from the text.
Therefore, this criterion prevents shortcut solutions and enables a more faithful evaluation of visual grounding.
The final criterion prevents confusing ICL demonstrations. 
If multiple characters exhibit similar behaviors across all demonstrations, it becomes difficult to determine which entity corresponds to the character described in the text, potentially making the task unsolvable even with strong visual grounding. 
By eliminating such cases, this criterion ensures that performance reflects visual grounding rather than ambiguity.

We note that data for each shot setting is constructed independently following the above rules, rather than smaller-shot demonstrations being subsets of larger-shot ones.
As a result, as discussed in Sec.~\ref{sec:benchmark} and illustrated in Fig.~\ref{fig:picl_shot}, higher-shot settings require reasoning over increasingly complex relationships among demonstrations, rather than simply providing more context, while lower-shot settings may be easier due to requiring fewer inferences.



\subsection{Robustness across Random Seeds}
\label{sec:appx:multiple_seeds}

We summarize the mean and standard deviation over three random seeds in Tab.~\ref{tab:multiple_seeds}. 
As shown, our proposed ProCoRe significantly outperforms the baselines, with performance gains that are substantially larger than the observed standard deviations. 
ICL ordering-based baselines (\eg, RICES and MMICES) are deterministic, as the demonstrations and their order are uniquely determined by their ranking criterion. 
Consequently, their results are identical across random seeds and exhibit no variance.

\begin{table*}[t]
\centering
\resizebox{\textwidth}{!}{
\begin{tabular}{llcccccc}
\toprule
\textbf{Category} & \textbf{Method}
& \multicolumn{3}{c}{\textbf{PICL}}
& \multicolumn{3}{c}{\textbf{TrueMICL}} \\
\cmidrule(lr){3-5}
\cmidrule(lr){6-8}
& 
& \textbf{2-shot} & \textbf{3-shot} & \textbf{4-shot}
& \textbf{2-shot} & \textbf{4-shot} & \textbf{8-shot} \\
\midrule

Off-the-shelf & -
& $24.97{\pm}1.55$
& $29.18{\pm}1.50$
& $29.37{\pm}1.52$
& $49.37{\pm}1.29$
& $55.58{\pm}2.60$
& $62.47{\pm}2.40$ \\
\midrule

\multirow{3}{*}{Retrieval \& Ordering}
& RICES
& $26.44{\pm}0.00$
& $27.14{\pm}0.00$
& $28.20{\pm}0.00$
& $47.16{\pm}0.00$
& $51.83{\pm}0.00$
& $61.86{\pm}0.00$ \\

& VICL
& $28.11{\pm}1.39$
& $26.10{\pm}1.75$
& $28.48{\pm}1.89$
& $50.06{\pm}1.59$
& $55.11{\pm}1.74$
& $61.66{\pm}0.53$ \\

& MMICES
& $21.70{\pm}0.00$
& $23.82{\pm}0.00$
& $24.79{\pm}0.00$
& $48.40{\pm}0.00$
& $58.07{\pm}0.00$
& $63.04{\pm}0.00$ \\
\midrule

\multirow{2}{*}{Task Vector}
& MTV
& $13.85{\pm}2.10$
& $15.09{\pm}1.84$
& $13.24{\pm}1.25$
& $42.94{\pm}1.78$
& $45.39{\pm}0.32$
& $55.55{\pm}0.81$ \\

& FV
& $25.86{\pm}1.92$
& $28.02{\pm}2.10$
& $29.62{\pm}2.26$
& $48.97{\pm}1.22$
& $56.24{\pm}1.52$
& $60.74{\pm}1.10$ \\
\midrule

\multirow{4}{*}{ICL-Data Training}
& MiMIC
& $27.53{\pm}2.18$
& $28.86{\pm}2.57$
& $31.60{\pm}2.58$
& $54.64{\pm}1.57$
& $54.80{\pm}2.27$
& $61.22{\pm}1.25$ \\

& STV
& $32.46{\pm}2.13$
& $33.62{\pm}1.88$
& $33.93{\pm}2.63$
& $50.20{\pm}1.55$
& $57.83{\pm}1.82$
& $64.03{\pm}1.54$ \\

& MT-ICL
& $32.78{\pm}1.82$
& $33.03{\pm}2.27$
& $33.75{\pm}2.25$
& $50.42{\pm}1.86$
& $58.01{\pm}1.78$
& $62.84{\pm}1.80$ \\

& DARA
& $25.35{\pm}2.27$
& $28.35{\pm}2.03$
& $29.78{\pm}1.86$
& $50.94{\pm}1.78$
& $56.24{\pm}1.02$
& $63.58{\pm}2.13$ \\
\midrule

\multirow{7}{*}{Mitigating Hallucination}
& VOLCANO
& $24.92{\pm}2.26$
& $27.42{\pm}2.13$
& $29.53{\pm}2.34$
& $50.90{\pm}1.64$
& $57.66{\pm}1.76$
& $63.46{\pm}2.22$ \\

& PAI
& $23.43{\pm}2.21$
& $25.01{\pm}2.68$
& $27.78{\pm}2.62$
& $48.05{\pm}1.73$
& $56.32{\pm}1.66$
& $59.97{\pm}1.26$ \\

& IBD
& $22.92{\pm}2.69$
& $23.74{\pm}1.21$
& $26.34{\pm}2.39$
& $50.47{\pm}1.74$
& $55.51{\pm}1.28$
& $62.62{\pm}2.10$ \\

& LRV
& $28.95{\pm}1.82$
& $30.28{\pm}2.36$
& $32.45{\pm}2.78$
& $51.99{\pm}1.90$
& $57.24{\pm}1.18$
& $63.74{\pm}1.91$ \\

& SPARC
& $26.36{\pm}2.29$
& $27.14{\pm}1.60$
& $33.20{\pm}2.66$
& $47.41{\pm}1.26$
& $57.13{\pm}1.71$
& $61.19{\pm}1.43$ \\

& LURE
& $25.30{\pm}2.15$
& $29.78{\pm}2.10$
& $30.40{\pm}2.24$
& $46.08{\pm}1.46$
& $53.08{\pm}1.56$
& $60.38{\pm}2.46$ \\

& \textbf{ProCoRe (Ours)}
& $\mathbf{36.11{\pm}1.59}$
& $\mathbf{36.88{\pm}1.31}$
& $\mathbf{36.80{\pm}1.74}$
& $\mathbf{56.07{\pm}1.87}$
& $\mathbf{62.56{\pm}1.77}$
& $\mathbf{68.04{\pm}0.92}$ \\

\bottomrule
\end{tabular}
}
\caption{\textbf{Comparison on PICL and TrueMICL across 3 seeds}. 
    Task Vector and ICL-Data Training baselines use the training and evaluation splits from the same dataset.
    }
\label{tab:multiple_seeds}
\end{table*}

\subsection{Algorithm of \ours}
\label{sec:appx:pseoducode}

Algorithm~\ref{alg:prop_reward} provides a pseudocode for the proposed \ours.

\begin{algorithm}[b]
\caption{\ours}
\label{alg:prop_reward}
\begin{algorithmic}[1]

\Statex \textbf{Input:} Contrastive pair data $(I_1,t_1),(I_2,t_2)$, MLLM $f_\theta$, sigmoid function $\sigma$, reward weights $\lambda_I,\lambda_T,\lambda_s$, proposition decomposer $\mathrm{PD}$, image-text similarity function $\mathrm{sim}_I$, entailment score function $\mathrm{entail}$

\State Generate captions for contrastive images using 
$f$ 
\begin{equation}
    \hat{t}_1 \gets f_\theta(I_1), \quad
\hat{t}_2 \gets f_\theta(I_2)
\end{equation}

\State Decompose captions into propositions
$(\hat{P}_1,\hat{P}_2,\hat{P}_s)
\gets
\mathrm{PD}(\hat{t}_1,\hat{t}_2)$

\State Compute image contrastive reward
\[
R_I^{(P)}
\gets
\sigma\!\Big(
\mathrm{sim}_I(I_1,\hat{P}_1)
-
\mathrm{sim}_I(I_1,\hat{P}_2)
+
\mathrm{sim}_I(I_2,\hat{P}_2)
-
\mathrm{sim}_I(I_2,\hat{P}_1)
\Big)
\]

\State Compute text contrastive reward
\[
R_T^{(P)}
\gets
\sigma\!\Big(
\mathrm{entail}(t_1,\hat{P}_1)
-
\mathrm{entail}(t_2,\hat{P}_1)
+
\mathrm{entail}(t_2,\hat{P}_2)
-
\mathrm{entail}(t_1,\hat{P}_2)
\Big)
\]

\State Compute shared-content reward
\[
R_s^{(P)}
\gets
\sigma\!\Big(
\mathrm{sim}_I(I_1,\hat{P}_s)
+
\mathrm{sim}_I(I_2,\hat{P}_s)
+
\mathrm{entail}(t_1,\hat{P}_s)
+
\mathrm{entail}(t_2,\hat{P}_s)
\Big)
\]

\State Combine rewards
$
R^{(P)}
\gets
\lambda_I R_I^{(P)}
+
\lambda_T R_T^{(P)}
+
\lambda_s R_s^{(P)}
$

\State \Return $R^{(P)}$

\end{algorithmic}
\end{algorithm}

\subsection{Qualitative Analysis of Visual Grounding}
\label{sec:appx:attention_quali}
\begin{figure}
    \centering   
    \includegraphics[width=\linewidth]{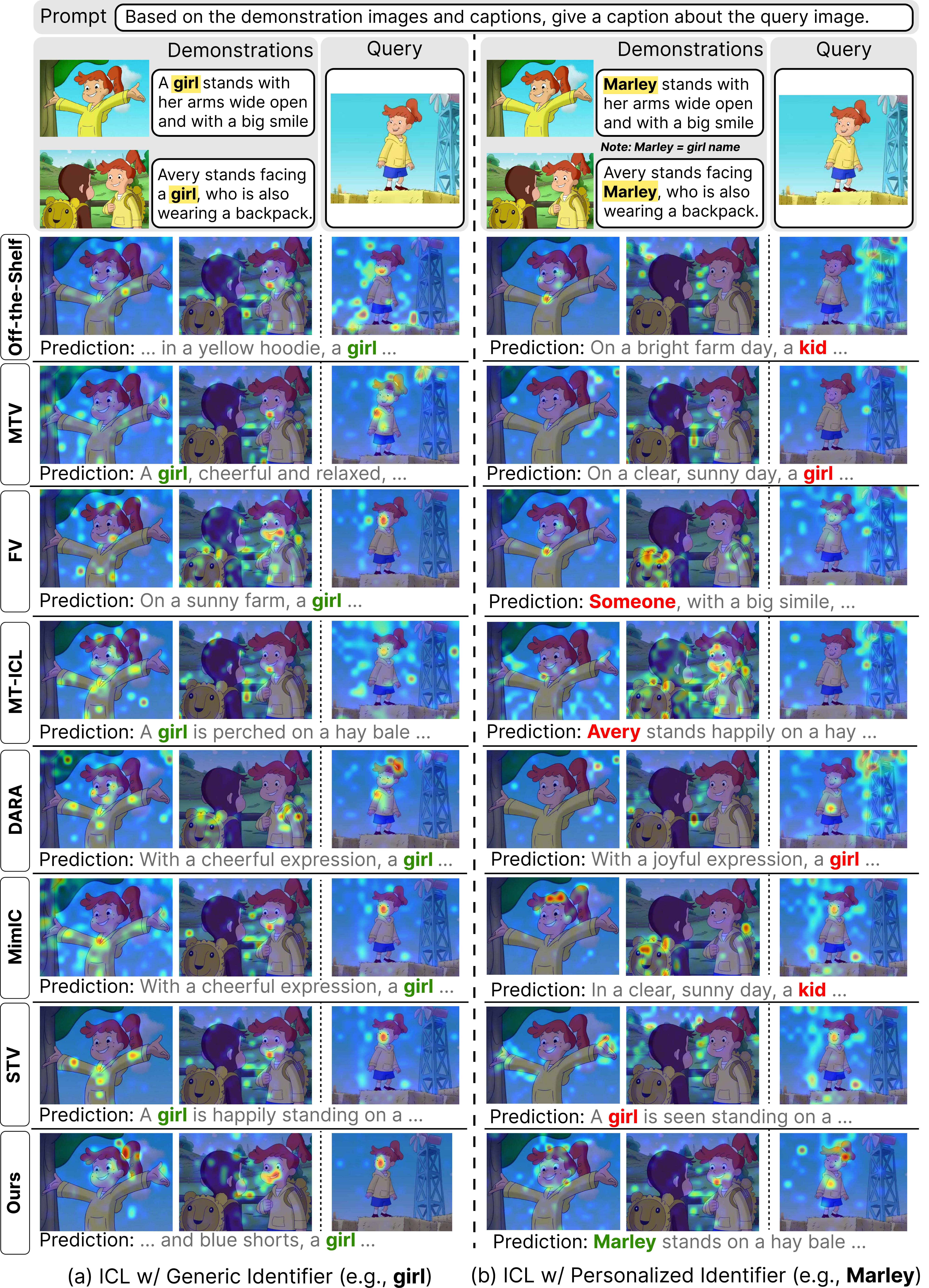}
    \vspace{-1em}
    \caption{
        \textbf{Visual grounding in multimodal ICL with generic vs. personalized identifiers.}
        Attention maps show visual-token attention when predicting the highlighted token using Qwen-VL-3-8B.
        (a) When a generic identifiers (\eg, ``{\setlength{\fboxsep}{0.5pt}%
\colorbox[HTML]{FFE32F}{\textbf{{girl}}}}'') seen during pre-training appears in the ICL demonstrations, both baselines and \ours attend to visual regions relevant to the predicted token (\ie, \textbf{\textcolor[HTML]{318C00}{girl}}).
        (b) In contrast, when the generic identifiers is replaced with a personalized identifier (\eg, ``{\setlength{\fboxsep}{0.5pt}%
        \colorbox[HTML]{FFE32F}{\textbf{{Marley}}}}''), baselines fail to attend to the relevant region and thereby fall back to generic terms (\eg, ``\textbf{\textcolor[HTML]{FF0004}{kid}}'', ``\textbf{\textcolor[HTML]{FF0004}{girl}}'' ``\textbf{\textcolor[HTML]{FF0004}{someone}}''), or incorrectly predict other names (\eg, ``\textbf{\textcolor[HTML]{FF0004}{Avery}}'' in MT-ICL), whereas \ours correctly predicts the personalized identifier (\ie, ``\textbf{\textcolor[HTML]{318C00}{Marley}}'') with strong visual grounding. 
    }
    \label{fig:quali_vg}
\end{figure}

We qualitatively compare visual grounding with ICL baselines in Fig.~\ref{fig:quali_vg}. 
As shown, when a generic concept appears in the demonstrations, both ICL baselines and \ours correctly predict the concept for the query image and attend to the corresponding visual regions when generating the token.
However, when the generic concept is replaced with a personalized identifier unseen during training, ICL baselines fail to attend to the relevant visual regions and instead output generic words (\ie, red words). 
This indicates that they do not effectively leverage visual demonstrations and rely on pre-trained semantic priors for prediction.
In contrast, \ours attends to the relevant visual regions in the query image and correctly predicts the personalized identifier (\ie, green word) introduced in the demonstrations, showing effective grounding and understanding of visual cues.


\subsection{Limitations and Future Work}
\label{sec:appx:limitations}

\ours improves visual grounding using standard multimodal contrastive benchmarks and shows consistent gains across datasets.
However, reliance on such curated benchmarks may limit scalability to larger-scale training and to data with more diverse contrastive visual attributes.
Combining with generative frameworks that can scalably produce high-quality multimodal contrastive data across diverse visual attributes (\eg, GenOL~\cite{seo2025genol}) could further enhance its scalability and real-world applicability.
Moreover, while \oursbenchmark evaluates visual grounding in multimodal ICL via captioning, real-world multimodal tasks extend beyond captioning to more complex reasoning.
Extending evaluation to such tasks would better reflect real-world settings.

\subsection{Impact Statements}
\label{sec:appx:impact_statements}

This work focuses on multimodal in-context learning, enabling MLLMs adapt to downstream tasks without parameter updates using a few demonstrations.
However, since \ours improves visual grounding via post-training with a contrastive reward,  there is a potential for the unintended amplification of issues such as model bias and ethical misalignment from bias in training data. 
We are committed to taking all necessary measures to address these risks, although tackling these concerns is not the primary focus of this work.


\clearpage









\end{document}